%% file: neurips_2026.tex
\documentclass{article}

\PassOptionsToPackage{numbers, compress}{natbib}
\usepackage[final,main]{neurips_2026}

\usepackage[utf8]{inputenc}
\usepackage[T1]{fontenc}
\usepackage[english]{babel}
\usepackage{hyperref}
\usepackage{url}
\usepackage{booktabs}
\usepackage{multirow}
\usepackage{makecell}
\usepackage{enumitem}
\usepackage{amsfonts}
\usepackage{amssymb}
\usepackage{amsmath}
\usepackage{amsthm}
\usepackage{mathtools}
\usepackage{nicefrac}
\usepackage{microtype}
\usepackage{xcolor}
\usepackage{algorithm}
\usepackage{algorithmic}
\usepackage{float}
\usepackage{tikz}
\usepackage{arydshln}
\usepackage{tcolorbox}
\tcbuselibrary{skins,breakable}
\usepackage{wrapfig}
\usepackage{graphicx}
\usepackage{subcaption}
\usepackage{epigraph}

\usetikzlibrary{arrows.meta, positioning, calc, fit, backgrounds}

\graphicspath{{./}{figures/}}

\definecolor{kept}{RGB}{46, 139, 87}
\definecolor{error}{RGB}{220, 53, 69}
\definecolor{discard}{RGB}{180, 180, 180}
\definecolor{success}{RGB}{34, 139, 34}
\definecolor{promptbg}{RGB}{248,248,248}
\definecolor{promptframe}{RGB}{180,180,180}

\providecommand{\E}{\mathbb{E}}

\newtcolorbox{promptbox}[1][]{
  colback=promptbg,
  colframe=promptframe,
  fonttitle=\bfseries\small,
  boxrule=0.5pt,
  arc=2pt,
  left=6pt,
  right=6pt,
  top=4pt,
  bottom=4pt,
  breakable,
  #1
}

\theoremstyle{plain}
\newtheorem{theorem}{Theorem}[section]

\theoremstyle{definition}

\theoremstyle{plain}
\newtheorem{example}[theorem]{Example}
\newtheorem{observation}[theorem]{Observation}
\newtheorem{remark}[theorem]{Remark}

\title{Iterative Policy Optimization: Batch-Online Multi-Turn RL via Single-Turn RL}

\author{%
  Daniel R.~Jiang\thanks{Equal contribution.} \\
  Meta \\
  \And
  Ankur Samanta\textsuperscript{*}\thanks{Work done at Meta.} \\
  Meta \\ Columbia University \\
  \And
  Yukai Yang \\
  Meta \\
  \And
  Jalaj Bhandari \\
  Meta \\
  \AND
  R\'emi Munos \\
  FAIR at Meta \\
  \And
  Tyler Lu \\
  Meta \\
}

\begin{document}

\maketitle

\begin{abstract}
\input{0_abstract}
\end{abstract}

\input{1_intro}
\input{2_formulation}

\input{4_1_reduction}
\input{4_2_exps}
\input{5_related}
\input{6_conclusion}

\begin{ack}
This work was supported by Meta. The authors declare no competing financial interests outside of their employment and affiliation with Meta and Columbia University.
\end{ack}

\bibliographystyle{plainnat}
\bibliography{refs}

\clearpage
\input{7_appendix}
\input{experiments_appendix}

\clearpage
\newpage
\input{checklist.tex}

\end{document}

%% file: 0_abstract.tex
Practical LLM agents often operate over multi-turn conversations where success is determined only after the full interaction ends. Most multi-turn RL methods train via on-policy rollouts, but unlike in single-turn RL, the policy cannot produce a trajectory alone, since an external environment must respond after each agent turn. For conversational agents, this environment is a user, but real users are generally unavailable inside the training loop and simulated users are difficult to build faithfully. Separately, real-world deployment is rarely fully online or fully offline. The common production pattern is called \emph{batch online}, where the current policy is deployed to collect a batch of interaction data, then retrained on that batch and redeployed.
We show that this batch-online setting creates an opportunity for applying the classical \emph{approximate policy iteration} algorithm. Our central observation is that running standard token-level, single-turn GRPO with a learned turn-level $Q$-function as the reward model is a policy improvement step for the multi-turn problem. Building on this, we present \emph{Iterative Policy Optimization} (IterPO), which alternates between fitting $Q^\pi$ from logged Monte Carlo returns via standard reward modeling (policy evaluation) and running single-turn GRPO against $Q^\pi$ (policy improvement). During policy improvement, the learned $Q^\pi$ scores candidate responses by their expected downstream return, avoiding the need for simulator rollouts of the remaining conversation. Overall, this allows us to do \emph{principled multi-turn RL using only single-turn RL methods} without an interactive environment or user simulator inside the training loop. We demonstrate the effectiveness of IterPO on six multi-turn negotiation environments.

%% file: 1_intro.tex
\section{Introduction}

\looseness-1 Many conversational AI agents are designed to drive specific \emph{outcomes}, including resolving a support issue, completing a task, or helping a customer reach a decision. These interactions play out over multiple turns, and whether the agent succeeded is often only apparent after the conversation ends. Earlier responses matter only through their downstream effects on later ones, so optimizing for outcomes is a long-horizon credit-assignment problem. Yet most widely used RLHF tooling targets \emph{single-turn} objectives, creating a gap between what we want to optimize and the tools available to do it.

Multi-turn RL is more challenging than single-turn RLHF because the policy cannot produce a trajectory by itself. After each agent response, an external environment must supply the next observation. For coding and web agents, real executable systems can supply these observations~\citep{jimenez2024swebench,zhou2023webarena,xie2024osworld}, but conversational agents need a \emph{user simulator}~\citep{yao2024taubench,barres2025tau2bench}, which is difficult to build and validate~\citep{li2025persona} and for many bespoke tasks simply unavailable. Most multi-turn RL methods extend on-policy optimizers such as GRPO and PPO to full trajectories~\citep{copet2025cwm,wei2025webagent,wang2025ragen,feng2025gigpo,li2025turnppo,wei2025reinforcing,ji2026treegrpo}, and thus they require an interactive environment inside the training loop. This requirement limits their applicability to tasks for which a reliable environment or user simulator is available.

\looseness-1 Separately, real-world deployment is rarely fully online or fully offline. The common regime is \emph{batch online}~\citep{dong2026batchonline}, in which the current policy is deployed, a batch of real interactions is logged over a deployment window (e.g., an A/B test lasting days or weeks), and the policy is retrained offline before redeployment. We show that this batch-online regime creates an opportunity for \emph{approximate policy iteration} \citep{bertsekas1996neuro}. The key observation is that a standard token-level RLHF optimizer, when given a turn-level $Q$-function as its reward model, acts as a policy improvement operator for the response-level multi-turn problem. In our proposed approach, we learn a state-action value function $Q^\pi$ from the logged interactions that predicts the expected downstream return of a candidate response under the environment dynamics and current policy. We then use $Q^\pi$ to score candidate responses \emph{without} rolling out the remainder of the conversation. Because each deployment round produces on-policy trajectories, the approximation of $Q^\pi$ can be obtained via standard supervised reward modeling on Monte Carlo returns and requires no temporal-difference learning. Thus, both policy evaluation and policy improvement can be implemented with \emph{off-the-shelf} single-turn RLHF tools, without bespoke multi-turn training infrastructure and without requiring an environment or user simulator inside the training loop.

In this work, we make the following contributions.
\begin{itemize}
    \setlength{\itemsep}{0.2em}
    \setlength{\topsep}{0.2em}
    \setlength{\parsep}{0pt}
    \setlength{\partopsep}{0pt}
  \item \looseness-1 \textbf{Multi-turn to single-turn RLHF reduction.} We give a formal reduction of multi-turn policy improvement to a sequence of standard single-turn, token-level RLHF problems, enabling direct use of off-the-shelf RLHF tools. We prove that standard single-turn GRPO with a learned turn-level state-action value function $Q^\pi$ as the reward model can be viewed as an approximate policy improvement step for the multi-turn problem (Theorem~\ref{thm:main}). The learned $Q^\pi$ supplies temporal credit assignment across conversational turns, while the single-turn GRPO finds the policy that approximately optimizes this quantity.
  \item \textbf{Approximate policy iteration without a simulator.} We propose \emph{Iterative Policy Optimization} (IterPO), a simple batch-online approximate policy iteration method that alternates between fitting $Q^\pi$ to Monte Carlo returns from logged conversations and improving the policy with standard GRPO, using $Q^\pi$ to score candidate responses. By estimating the expected downstream return of each candidate response from the current conversation state, $Q^\pi$ enables policy improvement without rolling out the remaining conversation in an environment simulator.
  \item \textbf{Multi-turn environments and empirical evaluation.} We adapt six existing negotiation datasets into a suite of synthetic multi-turn environments for LLM conversational agents and open-source the code. Across these environments, we show that IterPO outperforms single-turn and multi-turn baselines over multiple rounds of data collection and retraining.

\end{itemize}

%% file: 2_formulation.tex
\section{Preliminaries}
\label{sec:problem}

\subsection{Multi-Turn RLHF}

We study post-training for conversational agents in a \emph{multi-turn} setting where the interaction alternates between environment outputs $x_t$ and agent actions $a_t$ as follows.
\begin{equation*}
  x_1, a_1, x_2, a_2, \ldots
\end{equation*}
\looseness-1 In the standard setting, we can think of $x_t$ simply as the message sent by the user (or customer), while $a_t$ is the message sent by the agent. However, by allowing the definition of $x_t$ to be flexible, we can instantiate other variations of the problem where additional information besides the user's message may arrive. In general, $x_t$ is a generic environment output that bundles all of the information emitted between an agent's consecutive actions.

\looseness-1 We model this as an MDP $(\mathcal S, \mathcal A, P, P_0, R, \gamma)$, where $\mathcal S$ is the state space, $\mathcal A$ is the action space, $P$ is the transition kernel, $P_0$ is the initial state distribution, $R$ is the reward function, and $\gamma \in [0, 1)$ is the discount factor.

\noindent \textbf{State representation.} The state representation is the full history up to the current turn,
\begin{equation}
  s_t = (x_1, a_1, x_2, a_2, \ldots, x_t) \in \mathcal S,
\end{equation}
with initial state $s_1 = x_1$ drawn from $P_0$. A stochastic policy $\pi_\theta(a\,|\,s)$ maps states to action distributions.

\looseness-1 \noindent \textbf{Transition dynamics and termination.} After action $a_t$ in state $s_t$, the environment produces the next output
\begin{equation}
  x_{t+1} \sim P(\cdot \,|\, s_t, a_t).
\end{equation}
The MDP includes an absorbing terminal state $\bot$. After the agent takes $a_t$ and the environment produces $x_{t+1}$, the episode may terminate. Termination can depend on $x_{t+1}$ (e.g., through a terminal flag indicating ``no further messages'' or an outcome indicating successful resolution of the conversation). Once in $\bot$, the process remains there.

\textbf{Rewards and objective.} The reward can be any function of the state, action, and next environment output; we write $r_t = R(s_t, a_t, x_{t+1})$. Given a stochastic policy $\pi(a\,|\,s)$, the expected discounted return from state $s$ is
\begin{equation*}
  V^{\pi}(s) = \mathbb{E}\left[\sum_{t=1}^{\infty} \gamma^{t-1} \, r_t \;\biggl|\; s_1=s,\; a_t \sim \pi\right].
\end{equation*}
Under $s_1 \sim P_0$, the objective is $J(\pi)=\mathbb{E}_{s_1\sim P_0}[V^{\pi}(s_1)]$.

\vspace{0.5em}
\begin{example}[Outcome-driven multi-turn RL]
  \label{ex:ar}
  Here, we define the outcome-driven multi-turn conversational RL problem. Let the agent's message be $a_t$. Let $c_t$ denote the customer message at turn $t$, and let $o_t$ denote an \emph{outcome} signal (e.g., task completion, issue resolution, or purchase) that is revealed after the customer's message $c_t$ arrives. We set $x_t := (c_t, o_t)$, so that the interaction has the form $(c_1, o_1), a_1, (c_2, o_2), a_2,\ldots, (c_\tau, o_\tau)$. We suppose that termination occurs at the first outcome, i.e., the first time $\tau$ where $o_\tau=1$. There is a sparse outcome reward $R(s_t, a_t, x_{t+1})$ that is only nonzero when $o_{t+1}=1$.
  
\end{example}

\subsection{Batch-Online Deployment}
\label{sec:batch-online}

We consider a \emph{batch-online} deployment setting. The current policy $\pi_i$ is deployed and interacts with the environment (i.e., real users) to produce a batch of complete multi-turn trajectories with observed outcomes. The policy is then retrained offline on this batch before the next deployment round. This deploy-collect-retrain cycle repeats for $K$ rounds.

\begin{equation*}
  \underset{\text{deploy}}{\pi_0}
  \;\xrightarrow{\text{collect}}\;
  \mathcal{D}_0
  \;\xrightarrow{\text{train}}\;
  \underset{\text{deploy}}{\pi_1}
  \;\xrightarrow{\text{collect}}\;
  \mathcal{D}_1
  \;\xrightarrow{\text{train}}\;
  \cdots
  \;\xrightarrow{\text{train}}\;
  \pi_K.
\end{equation*}

The setting is ``online'' because the data-generating policy changes between rounds, and ``batch'' because each round trains on a fixed dataset rather than updating continuously. This matches the update pattern of classical batch-mode reinforcement learning and approximate dynamic programming~\citep{bertsekas1996neuro,tsitsiklis1996feature,ernst2005tree,lange2012batch}.

%% file: 4_1_reduction.tex
\begin{figure*}[!t]
  \hfill
  \includegraphics[width=1.05\linewidth]{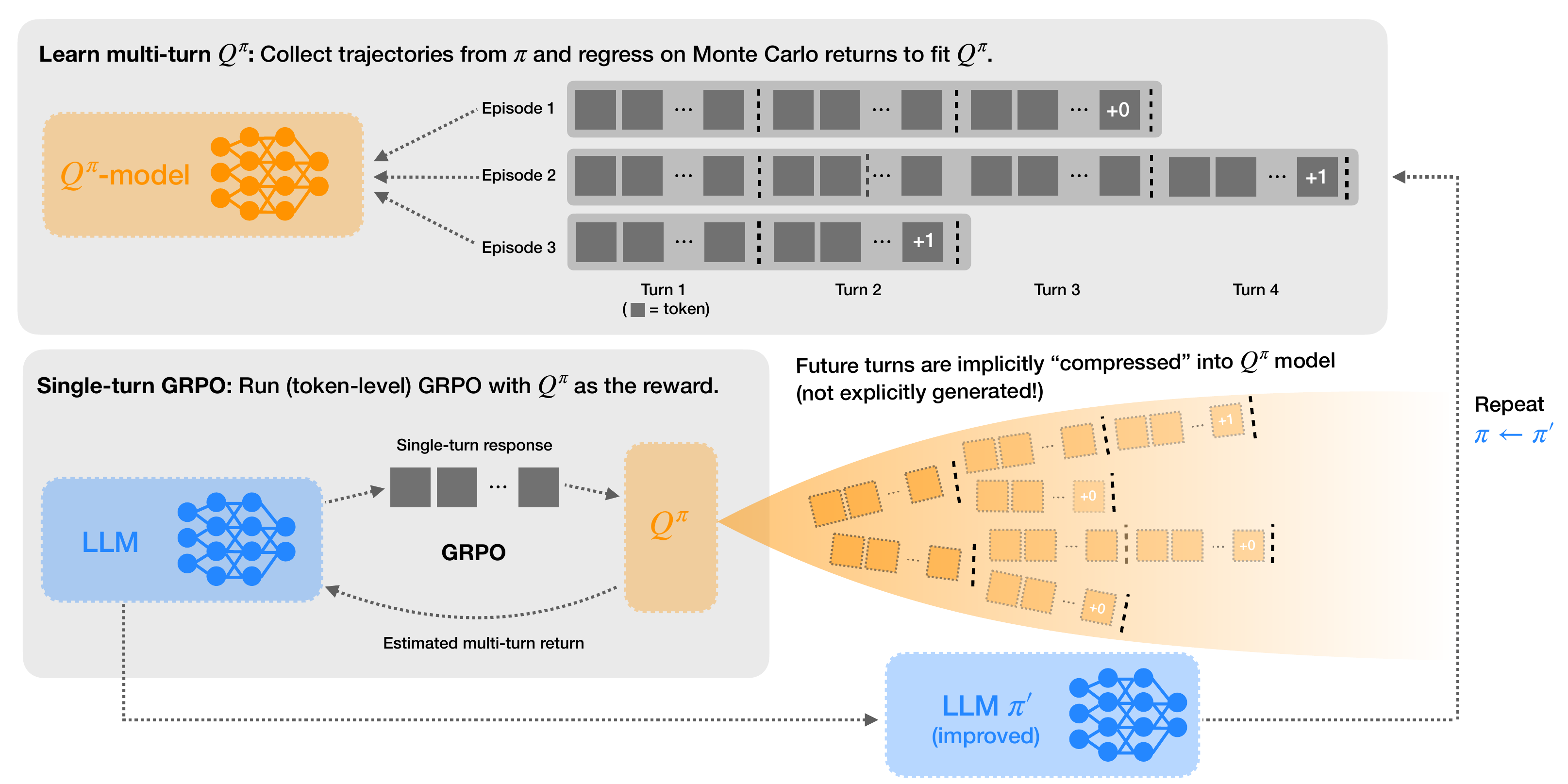}
  \caption{\textbf{Reducing multi-turn to single-turn RLHF via IterPO.} We visualize the main steps of our proposed approach. At the top, we collect multi-turn trajectories under the current policy $\pi$, compute Monte Carlo returns, and use standard reward-modeling procedures to fit the turn-level $Q$-function $Q^\pi$, which predicts the expected total multi-turn reward under $\pi$. At the bottom, holding $Q^\pi$ fixed, we run standard single-turn (token-level) GRPO using $Q^\pi$ as the reward model. Future conversation turns are implicitly ``compressed'' into $Q^\pi$, allowing candidate responses to be scored without simulator rollouts during policy improvement. The right side shows how iterating this procedure ($\pi \leftarrow \pi'$) yields multi-turn improvements using only single-turn RLHF tools. Since we proceed in a series of deployment rounds, we refer to this as ``batch-online.''}
  \label{fig:ippo}
\end{figure*}

\section{Multi-Turn to Single-Turn Reduction via IterPO}
\label{sec:reduction}

\looseness-1 In the single-turn setting, actions are typically defined at the \emph{token level},\footnote{There also exist response-level actions in a single-turn setting \citep{ahmadian2024back}, but in that case the problem is only a single stage, similar to a bandit problem.} with planning performed over the sequence of tokens that constitute a response  \citep{christiano2017deep, ziegler2019fine}. The state is the sequence of tokens generated so far, and the transition dynamics are to concatenate the latest action (token) with the previous state. In most instantiations, the overall response is then scored according to a reward model after the end-of-sequence (EOS) token. In contrast, the \emph{multi-turn} formulation naturally operates at the \emph{turn level}, where the atomic action is an agent's complete response to the previous turn. This allows for conversation-level planning that occurs over the sequence of responses \citep{li2016deep,wei2018airdialogue,shani2024multi}. Unlike token-level single-turn RLHF, where the transitions are deterministic token concatenations, the multi-turn setting involves unknown environment dynamics $P$ that govern how the conversation evolves.

We show that, under the batch-online setting defined in Section~\ref{sec:batch-online}, multi-turn RL can be cast as approximate policy iteration, with each step reducing to standard single-turn RLHF operations.

\subsection{The Reduction}

We now give two concrete observations that capture our \emph{policy iteration} perspective on multi-turn RL.

\begin{observation}[GRPO is an approximate policy improvement operator]
  \label{obs:imp}
  For a reward $R(s,a)$ over states $s$ and completions $a$, define $\mathtt{SingleTurnOpt}(R)$ to be a policy satisfying
  \begin{align*}
    \mathtt{SingleTurnOpt}(R)(\cdot\mid s) \in \arg\max_{\mu}\;
    \mathbb{E}_{a\sim\mu}\bigl[R(s,a)\bigr], \qquad \text{for every }s,
  \end{align*}
  where $\mu\in\mathcal{P}(\mathcal{A})$ is a distribution over completions. This idealized operator solves the single-turn reward maximization problem at every state. In practice, GRPO optimizes a shared token-level policy using the reward $R$, with updates averaged over sampled training states. GRPO can therefore be viewed as an approximation of $\mathtt{SingleTurnOpt}(R)$.

  Now suppose that $Q^\pi$, the state-action value function of a multi-turn policy $\pi$, is given to us. A policy that is greedy in the response-level action space with respect to $Q^\pi$ improves upon $\pi$ by the classical policy improvement theorem \citep{sutton1998reinforcement}. Our key insight is that since optimizing greedily in response space is difficult, we can instead run a single-turn RLHF optimizer with reward model $Q^\pi$, i.e.,
  \begin{equation}
    \pi' = \mathtt{SingleTurnOpt}(Q^\pi).
  \end{equation}
  By definition, $\pi'$ maximizes the expected $Q^\pi$-value at every state, so it scores at least as highly as the current policy $\pi$. The policy improvement theorem therefore gives $V^{\pi'}(s) \ge V^\pi(s)$ for every state $s$. Moreover, if we were given $Q^*$, then $\mathtt{SingleTurnOpt}(Q^*)$ would produce an optimal policy.\par
\end{observation}

\looseness-1 Notice that in order to guarantee policy $\pi'$ improves globally over $\pi$, we can relax the assumption that $\pi'$ is greedy with respect to $Q^\pi$ and only assume a local improvement property, as expressed in the following result (whose proof is given in Appendix~\ref{apx:proof.main}).

\begin{theorem}
  Assume we are able to obtain a (multi-turn) policy $\pi'$ which produces a single-turn improvement over $\pi$ from any state $s$, in the sense that \[\mathbb{E}_{a\sim \pi'(\cdot|s)}\left[ Q^{\pi}(s,a)\right] \geq \mathbb{E}_{a\sim \pi(\cdot|s)}\left[ Q^{\pi}(s,a)\right].\] Then the resulting (multi-turn) policy $\pi'$ is globally better than $\pi$, in the sense that $V^{\pi'}(s_0)\geq V^\pi(s_0)$ for any initial state $s_0$.
  \label{thm:main}
\end{theorem}

\begin{remark}[Approximate policy improvement]
  In practice, both $Q$-estimation and policy optimization are approximate. Appendix~\ref{apx:approx_kl} extends the idealized result to a learned $Q$-function with uniform estimation error $\epsilon_Q$ and per-state optimization error $\delta_s$.
\end{remark}

In practice, we approximate this policy improvement step using standard RLHF methods such as GRPO \citep{shao2024deepseekmath}, PPO \citep{schulman2017proximal}, or REINFORCE \citep{ahmadian2024back}. In this work, we use GRPO to implement the policy improvement step.

\begin{remark}[Token-level RL for response-level policy improvement]
  Observation~\ref{obs:imp} follows from the classical policy improvement theorem. Policy improvement provides a bridge between token-level and response-level RL because a token-level RL method can serve as a policy improvement operator for the response-level problem. Token-level RL methods have been shown to be empirically effective in single-turn settings, and Observation~\ref{obs:imp} lets us leverage them directly for multi-turn improvement.
\end{remark}

In Observation~\ref{obs:imp}, we assumed that we have access to $Q^\pi$. In our next observation, we show that approximating $Q^\pi$ can be achieved with standard supervised reward-model learning techniques used in single-turn RLHF.

\begin{observation}[Reward model fitting as policy evaluation]
  \label{obs:R}
  Consider the standard single-turn RLHF setup. Let $\mathcal D_\textnormal{single} = \{((s_i, a_i), r_i)\}_{i=1}^N$ be a supervised dataset of prompts $s_i$, responses $a_i$, and scalar targets $r_i$ (e.g., human labels). Reward modeling with scalar targets is a supervised learning problem.
  \begin{align*}
    &\mathtt{SingleTurnFitReward}(\mathcal D_\textnormal{single}) \in \mathop{\textnormal{argmin}}\limits_{R}\; \mathbb{E}_{((s,a),r)\sim \mathcal D_\textnormal{single}}\Bigl[\ell\bigl(R(s,a), r\bigr)\Bigr],
  \end{align*}
  \looseness-1 where $R$ belongs to some function class (e.g., a transformer) and $\ell$ is a supervised loss (e.g., mean squared error).

  Now consider a multi-turn policy $\pi$ and a dataset of $N$ on-policy trajectories $\{\tau^{(i)}\}_{i=1}^N$ collected by rolling out $\pi$. Each trajectory is of the form
  \[
    \tau^{(i)} = \bigl(s^{(i)}_1, a^{(i)}_1, r^{(i)}_1, s^{(i)}_2, a^{(i)}_2, r^{(i)}_2, \ldots, s^{(i)}_{T_i}\bigr),
  \]
  where $T_i$ is the (possibly varying) horizon of trajectory $i$. For each time step $t \in \{1,\ldots,T_i-1\}$, define the so-called Monte Carlo return to be
  \[
    G^{(i)}_t \;=\; \sum_{k=0}^{T_i-t-1} \gamma^k\, r^{(i)}_{t+k},
  \]
  where $\gamma \in [0,1)$ is the discount factor. We can then form a supervised dataset of state-action pairs and returns,
  \begin{equation}
    \mathcal D^{\pi}_{\textnormal{multi}} = \bigl\{\bigl((s^{(i)}_t, a^{(i)}_t), G^{(i)}_t\bigr) : i \in [N],\; t \in [T_i-1]\bigr\}.
    \label{eq:Dpi}
  \end{equation}
  \looseness-1 Monte Carlo policy evaluation \cite{sutton1998reinforcement} then takes the form
  \[
    Q^\pi \approx \mathtt{SingleTurnFitReward}(\mathcal D^\pi_{\mathrm{multi}}),
  \]
  a single-turn RLHF operation.\footnote{Why do we focus on $Q^\pi$ rather than aim for $Q^*$? This would likely involve an algorithm like fitted $Q$-iteration, but note that this requires the implementation of a max operator (over responses) to generate $Q$-labels. This could be another GRPO step, but the number of such iterations is inherently limited, since repeated policy improvement steps can quickly push the policy into regions of the response space that are out-of-distribution with respect to the original data. This is why we leverage the ability to do periodic data collection and perform $Q$-evaluation for the response-level problem.}
\end{observation}

\begin{remark}
  In Observation \ref{obs:R}, we made use of the Monte Carlo variant of policy evaluation in order to most easily leverage single-turn reward modeling tools. Other methods for performing policy evaluation include SARSA \cite{sutton1998reinforcement} and off-policy $Q$-evaluation \citep{sutton1998reinforcement,lagoudakis2003least}. These methods both use temporal differences (TD), which can have lower variance than Monte Carlo, but the downside is that TD-learning in an LLM setting is expensive and potentially unstable \citep{hong2024q}. In multi-turn settings with a reasonable horizon (i.e., the number of turns), Monte Carlo policy evaluation via standard reward modeling should suffice and has the added benefit of simplicity. However, if the data are collected using a policy other than $\pi_i$, then off-policy $Q$-evaluation would be required.
\end{remark}

\begin{remark}
Multi-turn RL can be formulated as a monolithic token-level optimization problem (e.g., via PPO). However, this is often impractical. Since environment transitions depend on external user responses, a token-level trainer needs to ``wait'' for customer responses. In addition, token-level methods may suffer from credit assignment challenges, as sparse terminal rewards must be propagated through thousands of tokens across multiple turns. Our reduction avoids these pitfalls by using a $Q$-function to compress the long-horizon dependency into a response-level reward, enabling the use of stable single-turn RLHF tools.
\end{remark}

\subsection{The IterPO Algorithm}

\looseness-1 At each IterPO round, we learn a turn-level $Q$-function $Q^\pi$ that estimates the (multi-turn) value of the current policy $\pi$ at the response level, and then run a single-turn, token-level RLHF method (in our case GRPO) using $Q^\pi$ as the reward model to obtain an improved policy $\pi'$. The justification for this algorithm comes directly from Observations~\ref{obs:imp} and \ref{obs:R}, which together constitute a single step of \emph{policy iteration}.

The pseudocode for IterPO is given in Algorithm \ref{alg:onlinebatch}. As we mentioned above, one of the main advantages of IterPO is that it can directly leverage mature and widely available off-the-shelf single-turn RLHF tools \citep{vonwerra2022trl,xFormers2022,sheng2024hybridflow}.

\begin{algorithm}
  \caption{IterPO for Multi-Turn RL}
  \begin{algorithmic}[1]
    \label{alg:onlinebatch}
    \STATE Initialize $\pi_0$ (e.g., prompt an instruction-tuned model to suggest a response given a conversation). Set number of deployment rounds $K$.\vspace{0.25em}
    \STATE Initialize $i \leftarrow 0$.\vspace{0.25em}
    \WHILE{$i < K$}
    \STATE Deploy $\pi_i$ and collect trajectories $\{\tau^{(i,j)}\}_{j=1}^N$.\vspace{0.25em}
    \STATE Create the dataset $\mathcal D^{\pi_i}_{\textnormal{multi}}$ according to \eqref{eq:Dpi}.\vspace{0.25em}
    \STATE Compute $Q^i \leftarrow \mathtt{SingleTurnFitReward}(\mathcal D^{\pi_i}_{\textnormal{multi}})$.\vspace{0.25em}
    \STATE Compute $\pi_{i+1} \leftarrow \mathtt{SingleTurnOpt}(Q^i).$\vspace{0.25em}
    \STATE Update $i \leftarrow i + 1$.\vspace{0.25em}
    \ENDWHILE
    \STATE \textbf{return} $\pi_K$.
  \end{algorithmic}
\end{algorithm}

Each round of Algorithm~\ref{alg:onlinebatch} consumes only the logged trajectories from the previous deployment, so no user simulator or interactive environment runs inside the training loop.

\subsection{Implementation via a Series of A/B Tests}
In practice, $\pi_0$ consists of simply prompting a base LLM to generate relevant suggestions based on the conversation session and other contextual information. Trajectories from $\pi_0$ can be collected by deploying it through an A/B test, from which we can learn an approximation to $Q^{\pi_0}$ (see top part of Figure \ref{fig:ippo}). After a policy improvement step (i.e., running GRPO on $Q^{\pi_0}$; see bottom part of Figure \ref{fig:ippo}) that results in a new policy $\pi_1$, we can deploy $\pi_1$ in an A/B test and track outcome metrics, such as the number of purchases. This A/B test for $\pi_1$ also serves to collect trajectories, from which we can repeat the above steps to arrive at $\pi_2$, and so on. Once we are satisfied with a particular policy $\pi_K$, it can be deployed to production.

\subsection{Action Coverage and Off-Policy Data Collection}
\looseness-1 One important consideration is the coverage of the action space of each iteration's collected dataset, which affects how well the learned $Q$-function generalizes across the action space. This can be an issue if, for example, $\pi_0(\cdot \,|\,s)$ is near deterministic and we use on-policy Monte Carlo policy evaluation. The resulting $Q^{\pi_0}(s,a)$ would therefore only be reliable at a small set of actions $a$ that are covered by $\pi_0(\cdot \,|\,s)$. To resolve this, we may need to consider collecting data using a behavioral policy $\pi_0'$ that is significantly different from $\pi_0$ in the sense that it is much more exploratory in its actions. For example, one way to do this is via \emph{persona exploration} where we can prompt the LLM to randomly select an array of characteristics such as tone, emoji usage, response length, etc. The trade-off with collecting data using $\pi_0'$ is that we must now perform an off-policy evaluation to learn $Q^{\pi_0}$ from trajectories collected from $\pi_0'$. However, this becomes more complex than on-policy Monte Carlo policy evaluation since it requires off-policy correction, for example through importance weighting or a TD-based evaluation procedure \citep{lagoudakis2003least}.

%% file: 4_2_exps.tex
\section{Numerical Experiments}
\label{sec:exps}

\subsection{Multi-Turn Environments}

We evaluate our method across six multi-turn negotiation environments that vary in the number of issues under negotiation, the information available to each agent, and the structure of the agreement space.  Each environment consists of two LLM-based agents (the trained agent $\pi$ and a static environment agent $\pi_{\text{env}}$) that alternate free-form natural-language turns until an agreement or breakdown is reached. As detailed in the environment-specific appendices, the environments use prior datasets and negotiation frameworks to define scenario variables and task structure, which we then use to generate new multi-turn LLM conversations. The environment parsers enforce task-specific agreement formats so malformed or one-sided deals are not incorrectly scored as valid agreements. Both agents are simulated with Qwen-2.5-7B-Instruct (Qwen7b). We briefly describe each environment below; full details including agent prompts and reward formulations are in the appendix.

\noindent \textbf{Craigslist Negotiation} (Appendix~\ref{appendix:craigslist}).
A single-issue price negotiation using the CraigslistBargain dataset \citep{he2018decoupling}. $\pi$ plays a vendor aiming to sell at favorable prices; $\pi_{\text{env}}$ plays a buyer with a private target price. Reward is the sale price divided by the listing price.

\noindent \textbf{Ultimatum Game} (Appendix~\ref{appendix:ultimatum}).
Two agents negotiate how to split a money pool \citep{bianchi2024negotiationarena}. $\pi$ is the proposer; $\pi_{\text{env}}$ is a responder who can accept, reject, or counter-propose. Reward is $\pi$'s share divided by half the pool size.

\noindent \textbf{Deal or No Deal} (Appendix~\ref{appendix:deal_or_no_deal}).
Two agents divide a shared pool of items (books, hats, balls) with private per-item values \citep{lewis2017deal}. Closure is bilateral. A partition is written as a \texttt{MINE}/\texttt{THEIRS} box and a deal closes only when the other party emits an explicit \texttt{ACCEPT}. Reward is points earned divided by $U_{\max}/2$ (5 for this dataset).

\noindent \textbf{CaSiNo} (Appendix~\ref{appendix:casino}).
Two campers divide packages of Food, Water, and Firewood, each with a private priority ordering over supply types \citep{chawla2021casino}. Closure follows the same bilateral \texttt{MINE}/\texttt{THEIRS} + \texttt{ACCEPT} protocol. Reward is points earned divided by $U_{\max}/2$ (18 for this dataset).

\looseness-1 \noindent \textbf{Job Interview} (Appendix~\ref{appendix:job_interview}).
A five-issue job offer negotiation with cross-issue dependencies and private utility functions, adapted from \citet{yamaguchi2021eacl}. Reward is a weighted sum of normalized per-issue scores.

\noindent \textbf{Alliance Negotiation} (Appendix~\ref{appendix:alliance}).
Two stakeholders negotiate over five project issues, each with a private score matrix and BATNA threshold, with issue structure drawn from \citet{abdelnabi2024llm}. Reward is the deal score divided by 50.

\subsection{Experimental Setup}
\label{sec:experimental_setup}

We simulate a realistic deployment setting in which a model is iteratively retrained and deployed over multiple rounds. The environment agent $\pi_{\text{env}}$ (an LLM) serves as a reproducible stand-in for real-world deployment. It generates each round's logged batch of conversations but is never queried inside the training loop, which operates purely on the logged trajectories.

In each round, the current policy is deployed to interact with the environment and collect a batch of trajectories; we then use this newly collected data to fine-tune the model before deploying it again. We use ``kernel'' to refer to one fixed environment instance, including the item values, private preferences, thresholds, and other sampled variables that define a negotiation scenario. Each round generates 400 conversations (100 distinct in-training kernels, each simulated 4 times independently) over $K=4$ rounds total. We report in-training performance on the 100 in-training kernels from each deployment round and held-out performance on 100 held-out kernels per environment. The per-turn discount factor is $\gamma = 0.9$, which encourages the agent to reach favorable outcomes efficiently. For each environment, we report the expected discounted return $J(\pi)$. Since our environments only produce a terminal reward at the end of the conversation, the total reward of an episode of length $\tau$ is $\gamma^{\tau-1}\, r_\tau$. This captures both the quality of the negotiated outcome and the efficiency with which it is reached. Results are averaged over 3 seeds. All algorithms are evaluated under the same iterative deployment protocol. We compare the following:
\begin{itemize}
  \item \textbf{IterPO}: Our proposed approach (Algorithm~\ref{alg:onlinebatch}), which alternates across deployment rounds between fitting a turn-level $Q$-function to Monte Carlo returns from logged conversations and using that $Q$-function as the reward model for single-turn, token-level GRPO.
  \item \textbf{ST-GRPO}: A myopic baseline that applies GRPO to each individual turn, treating them as independent decision problems ($\gamma = 0$).
  \item \textbf{MT-SFT}: A supervised baseline that fine-tunes the model to imitate the agent's own messages from successful trajectories (those ending in agreement), with the loss masked to the agent's turns. It performs no explicit reward optimization.
  \item \textbf{MT-KTO}: A multi-turn version of KTO~\citep{ethayarajh2024kto} applied to the agent's messages from each collected conversation, using agreement and no-agreement outcomes as binary feedback.
  \item \looseness-1 \textbf{MT-PPO}: A PPO baseline run under the same multi-turn batch-online experimental protocol as the other methods where applicable, using a learned value head for the logged round batch.
  \item \textbf{ArCHer} \citep{zhou2024archer}: A hierarchical actor-critic method for multi-turn RL that learns response-level $Q$- and $V$-functions. We use its offline variant (an implicit-$Q$-learning critic with an advantage-weighted-regression actor), which is suited to our batch-online data regime. The original implementation uses a RoBERTa-based critic~\citep{liu2019roberta}.
  \item \textbf{ArCHer-LC} (LLM-Critic): A variant of ArCHer whose critic is built on a frozen copy of the policy's language model with lightweight value heads, the same critic architecture as IterPO's $Q$-function.
\end{itemize}
Full experimental details are provided in Appendix~\ref{appendix:hyperparameters}.

\subsection{Results}
\label{sec:results}

\begin{figure}[t]
    \centering
    \includegraphics[width=0.89\linewidth]{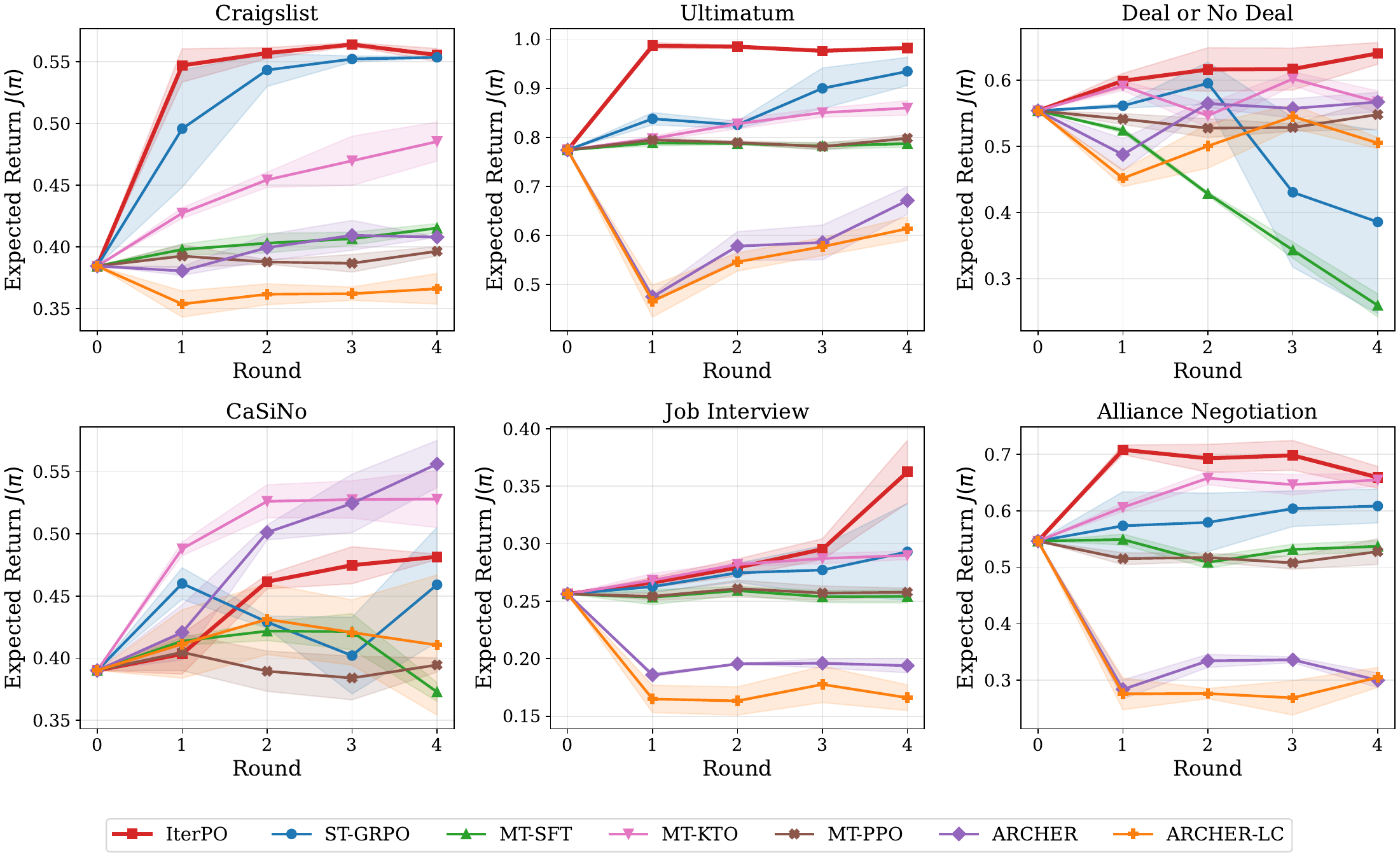}
    \caption{In-training expected discounted return $J(\pi)$ for Qwen7b across the $K{=}4$ deployment rounds. Error bands show $\pm1$ SE across 3 seeds. Held-out return at the checkpoint with the best in-training return is reported in Table~\ref{tab:summary}.}
    \label{fig:composite_expected_return}
\end{figure}

Table~\ref{tab:summary} reports expected discounted return on the held-out kernels. Figure~\ref{fig:composite_expected_return} reports the same quantity in-training across all deployment rounds; Table~\ref{tab:summary_online} gives the corresponding in-training values. We find that IterPO improves across multiple deployment rounds in most environments and does not exhibit reward collapse. On held-out kernels, IterPO achieves the highest expected discounted return in five of six environments and improves over the initial policy in every case. The single-turn ablation improves over the initial policy in some environments but falls short of IterPO, suggesting that multi-turn credit assignment matters in this setting. MT-SFT only weakly improves over the initial policy, while MT-KTO and MT-PPO are stronger baselines but do not match IterPO overall.

\looseness-1 In some settings, agreement rates approach their ceiling early. In Craigslist and Alliance Negotiation, agreement rates rise above 95\% after the first training round, while Ultimatum starts near this regime and remains there throughout training (Table~\ref{tab:agreement_rate_rounds}). With this sparse terminal reward, once agreement frequency saturates, later rounds have less room to improve and return depends more on the quality of the final deal, so performance can plateau or slightly decline. We therefore evaluate each trained method at the checkpoint with the best in-training return over the $K=4$ deployment rounds. This matches the deployment protocol. After each round, the practitioner can keep the best policy observed so far instead of using the final checkpoint by default.

\begin{table*}[ht]
\centering
\caption{Expected discounted return $J(\pi)$ on held-out kernels. ``Base'' reports the initial instruction-tuned policy; trained methods are evaluated at the checkpoint with the best in-training return. Means $\pm 1$ SE across 3 seeds.\protect\footnotemark}
\label{tab:summary}
\small
\setlength{\tabcolsep}{4pt}
\begin{tabular}{lcccccc}
\toprule
Method & Craigslist & Ultimatum & DealOrNoDeal & CaSiNo & Job Interview & Alliance \\
\midrule
Base & $0.36$ & $0.80$ & $0.52$ & $0.38$ & $0.25$ & $0.48$ \\
\midrule
MT-SFT & $0.41{\scriptstyle\pm0.01}$ & $0.79{\scriptstyle\pm0.01}$ & $0.50{\scriptstyle\pm0.05}$ & $0.42{\scriptstyle\pm0.01}$ & $0.25{\scriptstyle\pm0.00}$ & $0.51{\scriptstyle\pm0.02}$ \\
MT-KTO & $0.48{\scriptstyle\pm0.01}$ & $0.86{\scriptstyle\pm0.01}$ & $0.56{\scriptstyle\pm0.00}$ & $\boldsymbol{0.54{\scriptstyle\pm0.01}}$ & $0.29{\scriptstyle\pm0.01}$ & $0.59{\scriptstyle\pm0.00}$ \\
MT-PPO & $0.38{\scriptstyle\pm0.02}$ & $0.78{\scriptstyle\pm0.01}$ & $0.51{\scriptstyle\pm0.03}$ & $0.38{\scriptstyle\pm0.02}$ & $0.24{\scriptstyle\pm0.01}$ & $0.51{\scriptstyle\pm0.05}$ \\
ArCHer & $0.45{\scriptstyle\pm0.01}$ & $0.91{\scriptstyle\pm0.03}$ & $0.60{\scriptstyle\pm0.03}$ & $0.53{\scriptstyle\pm0.03}$ & $0.26{\scriptstyle\pm0.01}$ & $0.57{\scriptstyle\pm0.01}$ \\
ArCHer-LC & $0.38{\scriptstyle\pm0.01}$ & $0.86{\scriptstyle\pm0.02}$ & $0.53{\scriptstyle\pm0.04}$ & $0.40{\scriptstyle\pm0.04}$ & $0.25{\scriptstyle\pm0.01}$ & $0.49{\scriptstyle\pm0.06}$ \\
ST-GRPO & $0.54{\scriptstyle\pm0.00}$ & $0.94{\scriptstyle\pm0.05}$ & $0.58{\scriptstyle\pm0.06}$ & $0.46{\scriptstyle\pm0.03}$ & $0.28{\scriptstyle\pm0.04}$ & $0.55{\scriptstyle\pm0.02}$ \\
\midrule
IterPO & $\boldsymbol{0.55{\scriptstyle\pm0.01}}$ & $\boldsymbol{0.99{\scriptstyle\pm0.00}}$ & $\boldsymbol{0.70{\scriptstyle\pm0.01}}$ & $0.45{\scriptstyle\pm0.03}$ & $\boldsymbol{0.35{\scriptstyle\pm0.03}}$ & $\boldsymbol{0.65{\scriptstyle\pm0.01}}$ \\
\bottomrule
\end{tabular}
\end{table*}
\footnotetext{Full environment names: Craigslist Negotiation, Ultimatum Game, Deal or No Deal, CaSiNo, Job Interview, Alliance Negotiation (left to right).}

We also analyze where IterPO's improvement in discounted return comes from. Because the objective discounts by turn, shorter successful conversations directly improve return. IterPO-trained policies do shorten conversations while maintaining or improving agreement rates (Figure~\ref{fig:composite_conv_length}), but their undiscounted terminal rewards also improve in all six environments (Figure~\ref{fig:composite_terminal_reward}). The gains therefore reflect both terms in the discounted objective. The policy reaches acceptable agreements faster and reaches better final deals.

\paragraph{Behavioral patterns.}
We next analyze IterPO's conversation traces to see what behaviors emerge when training against a sparse outcome reward, without supervising or guiding the negotiation strategy itself (details in Appendix~\ref{appendix:behavioral}). In Craigslist, Ultimatum, CaSiNo, and Alliance Negotiation, the main behavioral change is that IterPO reaches acceptable proposals faster. The agent asks for the partner's priorities, proposes a complete agreement earlier, and spends fewer turns on score calculations or intermediate concessions. In Deal or No Deal and Job Interview, the same training also produces richer multi-issue bargaining behavior. In particular, IterPO policies reveal private values less often and use partner-stated information more often. They also trade concessions on lower-value issues for gains on higher-value ones, a strategy known as logrolling. In Job Interview, they concede on the contested issue while holding the rest of the package fixed, a form of selective accommodation \citep{pruitt1981negotiation, raiffa1982art}, showing that sparse outcome rewards can lead the policy to recognizable multi-issue bargaining strategies in these environments.

%% file: 5_related.tex
\section{Related Work}
\label{sec:related}

\paragraph{On-policy multi-turn RL.}
Most multi-turn RL for LLMs extends \emph{on-policy} policy-gradient methods to operate over entire trajectories. Multi-turn variants of GRPO and PPO have been applied to coding~\citep{copet2025cwm}, web~\citep{wei2025webagent}, and reasoning~\citep{wang2025ragen} agents, and a line of work refines turn-level credit assignment by branching from intermediate states or grouping repeated states~\citep{feng2025gigpo,li2025turnppo,wei2025reinforcing,ji2026treegrpo,zhu2026gagpo}. These methods are on-policy by construction, since their advantages are estimated from trajectories sampled by the current policy, so each update requires fresh rollouts. Our method instead consumes \emph{logged} on-policy trajectories collected between deployment rounds, scoring new candidate responses without rolling out subsequent conversation turns.

\paragraph{Multi-turn preference optimization.}

Another line of work learns from \emph{trajectory-level} preferences over entire conversations. MTPO~\citep{shani2024multi} uses mirror descent for online preference collection, \citet{xiong2024building} propose an iterative multi-turn extension of DPO~\citep{rafailov2023direct}, and \citet{shi2024direct} target the offline setting. Collecting pairwise preferences over long conversations is often impractical, and these methods require bespoke multi-turn algorithms. Our approach avoids both issues by reducing to standard single-turn techniques and operates in a batch-online regime closest to \citet{xiong2024building}.

\paragraph{Hierarchical multi-turn RL.}

The most closely related work is ArCHer~\citep{zhou2024archer}, which also distinguishes response-level and token-level problems and fits a $Q$-function to guide token-level optimization. ArCHer uses off-policy Bellman updates via IQL~\citep{kostrikov2022offline} and trains multiple critic networks ($Q$, $V$, and baseline), requiring careful tuning for stability. Our approach differs in two ways. First, we show that the token-level optimization step is a policy improvement step for the multi-turn problem (Thms.~\ref{thm:main} and \ref{thm:app}), which guarantees monotone improvement under exact $Q$-estimation and statewise policy improvement. Second, our algorithm requires only two phases (supervised $Q$-fitting followed by GRPO) with no TD learning. POAD~\citep{wen2024reinforcing} decomposes actions into token groups for PPO. REFUEL~\citep{gao2025refuel} simplifies ArCHer via regression but requires pairwise trajectory rollouts per state. \citet{nam2025efficient} define an abstract MDP over high-level tutoring actions, a different form of hierarchical decomposition.

\paragraph{Other offline approaches.}

\looseness-1 Several works optimize dialogue from static datasets~\citep{jaques2020human, jang2022gpt, snell2022offline, verma2022chai, chen2025broaden}. \citet{chen2025broaden} add inference-time search on top of an offline-trained model. Our method differs in that it supports periodic online updates, which enable approximate policy iteration across deployment rounds.

%% file: 6_conclusion.tex
\section{Limitations and Concluding Remarks}
\label{sec:conclusion}
\looseness-1 We present IterPO, a batch-online \emph{approximate policy iteration} algorithm for multi-turn RLHF that learns from logged on-policy trajectories without a user simulator inside the training loop. The method reduces multi-turn RL to a sequence of single-turn RLHF problems, namely reward modeling and GRPO, while reusing off-the-shelf single-turn tools with no bespoke multi-turn infrastructure. Empirically, we evaluate IterPO on six multi-turn negotiation environments and show that it outperforms both single-turn and multi-turn baselines in five of six settings. A key assumption underlying our reduction is that accurate $Q$-function estimation is possible via Monte Carlo returns, which could be more challenging in environments with very long horizons or high stochasticity. In principle, TD-based methods could improve $Q$-estimation in such settings, though they introduce additional complexities. In the environments we considered, TD-learning was not required to obtain good performance. What was needed, however, was a model whose initial negotiation tendencies produced enough agreements to provide a useful learning signal. If the initial post-trained policy almost always walks away (zero reward signal), additional prompting or SFT warm start may be needed.

\clearpage


%% file: 7_appendix.tex
\appendix
\label{A:A}

\section*{Contents of the Appendix}
\noindent
{\small
\setlength{\parskip}{0pt}
\begin{itemize}[leftmargin=2.5em,itemsep=1pt,topsep=2pt]
  \item[\ref*{apx:llm_usage}] \hyperref[apx:llm_usage]{LLM Usage}
  \item[\ref*{apx:proof.main}] \hyperref[apx:proof.main]{Proof of Theorem~\ref{thm:main}}
  \item[\ref*{apx:approx_kl}] \hyperref[apx:approx_kl]{Approximate KL-Regularized Policy Improvement}
  \item[\ref*{appendix:environments}] \hyperref[appendix:environments]{Environment Details}
    \begin{itemize}[leftmargin=1.75em,itemsep=0pt,topsep=1pt]
      \item[\ref*{appendix:craigslist}] \hyperref[appendix:craigslist]{Craigslist Negotiation}
      \item[\ref*{appendix:ultimatum}] \hyperref[appendix:ultimatum]{Ultimatum Game}
      \item[\ref*{appendix:deal_or_no_deal}] \hyperref[appendix:deal_or_no_deal]{Deal or No Deal}
      \item[\ref*{appendix:casino}] \hyperref[appendix:casino]{CaSiNo}
      \item[\ref*{appendix:job_interview}] \hyperref[appendix:job_interview]{Job Interview}
      \item[\ref*{appendix:alliance}] \hyperref[appendix:alliance]{Alliance Negotiation}
    \end{itemize}
  \item[\ref*{appendix:hyperparameters}] \hyperref[appendix:hyperparameters]{Training Details}
  \item[\ref*{appendix:additional_metrics}] \hyperref[appendix:additional_metrics]{Additional Metrics}
  \item[\ref*{appendix:online_return}] \hyperref[appendix:online_return]{In-Training Return}
  \item[\ref*{appendix:held_out_test}] \hyperref[appendix:held_out_test]{Held-Out Kernels}
  \item[\ref*{appendix:llama3b}] \hyperref[appendix:llama3b]{Additional Backbone: Llama3b}
  \item[\ref*{appendix:behavioral}] \hyperref[appendix:behavioral]{Systematic Behavioral Analysis}
\end{itemize}
}
\vspace{1em}

\section{LLM Usage}\label{apx:llm_usage}
LLMs serve as the trained policy models and fixed environment agents in our multi-turn experiments (Section~\ref{sec:exps}, Appendices~\ref{appendix:environments}, \ref{appendix:hyperparameters}, and~\ref{appendix:llama3b}), and as an automated judge for the behavioral analysis of negotiation transcripts (Section~\ref{sec:results} and Appendix~\ref{appendix:behavioral}). LLMs were also used to edit the manuscript for clarity.\par

\section{Proof of Theorem~\ref{thm:main}}\label{apx:proof.main}
\begin{proof}
By the performance difference lemma (see, e.g., Section~1.5 of \cite{agarwal2019reinforcement}), we have
\begin{align*}
    V^{\pi'}(s_0) - V^{\pi}(s_0) 
    &= \frac{1}{1-\gamma} \E_{s\sim d^{\pi'}_{s_0}} \, \E_{a\sim\pi'(\cdot|s) } \bigl[ A^\pi(s,a) \bigr]\\
    &= \frac{1}{1-\gamma} \E_{s\sim d^{\pi'}_{s_0}} \, \left[ \E_{a\sim\pi'(\cdot|s) } \bigl[ Q^\pi(s,a) \bigr] 
    - \E_{a\sim\pi(\cdot|s) } \bigl[ Q^\pi(s,a) \bigr]\right] \geq 0,
\end{align*}
where 
\begin{equation}
\label{eq:pdd}
    d^{\pi'}_{s_0}(s) = (1-\gamma) \sum_{t=0}^\infty \gamma^t \, \mathbb{P}(s_t = s \, | \, s_0, \pi') \; 
\end{equation}
is the discounted state visitation distribution under $\pi'$.
\end{proof}

\section{Approximate KL-Regularized Policy Improvement}\label{apx:approx_kl}
We extend Theorem~\ref{thm:main} to a KL-regularized reward-maximization objective. The extension allows the learned $Q$-function to have uniform estimation error $\epsilon_Q$ and the policy optimizer to incur state-dependent error $\delta_s$.\par
\begin{theorem}
    Let $\pi$ be a multi-turn policy, and let $\widehat{Q}$ approximate $Q^\pi$ with
    \[
        \|\widehat{Q}-Q^\pi\|_\infty \le \epsilon_Q.
    \]
    For each state $s$, define
    \[
        \mathcal L_s(\mu)
        =
        \E_{a\sim\mu(\cdot\,|\,s)}
        \bigl[\widehat{Q}(s,a)\bigr]
        -
        \beta\,\mathrm{KL}
        \bigl(\mu(\cdot\,|\,s)\,\|\,\pi(\cdot\,|\,s)\bigr),
    \]
    where $\beta\ge0$. Suppose $\pi'$ satisfies
    \[
        \mathcal L_s(\pi')
        \ge
        \sup_\mu \mathcal L_s(\mu)-\delta_s
    \]
    for every state $s$. Then, for any initial state $s_0$,
    \[
        V^{\pi'}(s_0)-V^\pi(s_0)
        \ge
        \frac{1}{1-\gamma}
        \E_{s\sim d^{\pi'}_{s_0}}
        \left[
            \beta\,\mathrm{KL}
            \bigl(\pi'(\cdot\,|\,s)\,\|\,\pi(\cdot\,|\,s)\bigr)
            -\delta_s-2\epsilon_Q
        \right].
    \]
    Consequently, $\pi'$ improves on $\pi$ whenever the expectation on the right-hand side is nonnegative.
    \label{thm:app}
\end{theorem}
The experiments use the unregularized case $\beta=0$.

\begin{proof}
Approximate optimality and feasibility of $\pi$ imply
\begin{align}
\mathcal L_s(\pi')
&\ge \sup_\mu \mathcal L_s(\mu)-\delta_s \nonumber\\
&\ge \mathcal L_s(\pi)-\delta_s \nonumber\\
&= \E_{a\sim\pi(\cdot\,|\,s)}
   \bigl[\widehat{Q}(s,a)\bigr]-\delta_s,
\label{eq:kl_reg_pi_with_errors}
\end{align}
where the last equality uses
$\mathrm{KL}(\pi(\cdot\,|\,s)\,\|\,\pi(\cdot\,|\,s))=0$.
Rearranging Equation~\eqref{eq:kl_reg_pi_with_errors} and applying
$|\widehat{Q}(s,a)-Q^\pi(s,a)|\le\epsilon_Q$ to both policy expectations gives
\begin{align}
&\E_{a\sim\pi'(\cdot\,|\,s)}\bigl[Q^\pi(s,a)\bigr]
-
\E_{a\sim\pi(\cdot\,|\,s)}\bigl[Q^\pi(s,a)\bigr] \nonumber\\
&\qquad\ge
\beta\,\mathrm{KL}
\bigl(\pi'(\cdot\,|\,s)\,\|\,\pi(\cdot\,|\,s)\bigr)
-\delta_s-2\epsilon_Q.
\label{eq:perf_diff_bd}
\end{align}
Because $\E_{a\sim\pi(\cdot\,|\,s)}[Q^\pi(s,a)]=V^\pi(s)$,
Equation~\eqref{eq:perf_diff_bd} implies

\begin{equation}
    \E_{a\sim\pi'(\cdot\,|\,s)} \bigl[ {A}^\pi(s,a)  \bigr] \ge {\beta}  \, \mathrm{KL}\bigl(\pi'(\cdot\,|\,s)\, \|\, \pi(\cdot\,|\,s) \bigr) - \delta_s - 2\epsilon_Q.
\end{equation}

Applying the performance difference lemma yields
\begin{align*}
V^{\pi'}(s_0)-V^\pi(s_0)
&=
\frac{1}{1-\gamma}
\E_{s\sim d^{\pi'}_{s_0}}
\E_{a\sim\pi'(\cdot|s)}
\bigl[A^\pi(s,a)\bigr]\\
&\ge
\frac{1}{1-\gamma}
\E_{s\sim d^{\pi'}_{s_0}}
\left[
\beta\,\mathrm{KL}
\bigl(\pi'(\cdot\,|\,s)\,\|\,\pi(\cdot\,|\,s)\bigr)
-\delta_s-2\epsilon_Q
\right].
\end{align*}
Thus, $\pi'$ weakly improves on $\pi$ whenever
\[
\beta\,
\E_{s\sim d^{\pi'}_{s_0}}
\left[
\mathrm{KL}\bigl(\pi'(\cdot\,|\,s)\,\|\,\pi(\cdot\,|\,s)\bigr)
\right]
\ge
\E_{s\sim d^{\pi'}_{s_0}}
\left[\delta_s+2\epsilon_Q\right],
\]
and the improvement is strict when the inequality is strict.

\end{proof}

%% file: experiments_appendix.tex
\section{Environment Details}
\label{appendix:environments}

\subsection{Craigslist Negotiation}
\label{appendix:craigslist}

\input{appendix_craigslist}

\subsection{Ultimatum Game}
\label{appendix:ultimatum}

\input{appendix_ultimatum}

\subsection{Deal or No Deal}
\label{appendix:deal_or_no_deal}

\input{appendix_deal_or_no_deal}

\subsection{CaSiNo}
\label{appendix:casino}

\input{appendix_casino}

\subsection{Job Interview}
\label{appendix:job_interview}

\input{appendix_job_interview}

\subsection{Alliance Negotiation}
\label{appendix:alliance}

\input{appendix_alliance}

\section{Training Details}
\label{appendix:hyperparameters}

\input{appendix_hyperparameters}

\section{Additional Metrics}
\label{appendix:additional_metrics}

\input{appendix_additional_metrics}

\section{In-Training Return}
\label{appendix:online_return}

Table~\ref{tab:summary} reports held-out return at the checkpoint selected by in-training performance. Here, we report in-training return on the 400 rollouts collected during each deployment round (100 in-training kernels $\times$ 4 conversations per kernel). Figure~\ref{fig:composite_expected_return} plots this quantity across rounds. Table~\ref{tab:summary_online} reports each method's best in-training round, while Table~\ref{tab:summary_r4} reports the final round.\par

\begin{table*}[ht]
\centering
\caption{Qwen7b in-training expected discounted return $J(\pi)$ at each method's best round. ``Base'' reports the initial instruction-tuned policy evaluated before training. Means $\pm 1$ SE across 3 seeds.}
\label{tab:summary_online}
\small
\setlength{\tabcolsep}{4pt}
\begin{tabular}{lcccccc}
\toprule
Method & Craigslist & Ultimatum & DealOrNoDeal & CaSiNo & Job Interview & Alliance \\
\midrule
Base & $0.38$ & $0.77$ & $0.55$ & $0.39$ & $0.26$ & $0.55$ \\
\midrule
MT-SFT & $0.42{\scriptstyle\pm0.00}$ & $0.79{\scriptstyle\pm0.01}$ & $0.52{\scriptstyle\pm0.00}$ & $0.42{\scriptstyle\pm0.01}$ & $0.26{\scriptstyle\pm0.00}$ & $0.55{\scriptstyle\pm0.01}$ \\
MT-KTO & $0.49{\scriptstyle\pm0.02}$ & $0.86{\scriptstyle\pm0.01}$ & $0.60{\scriptstyle\pm0.01}$ & $0.53{\scriptstyle\pm0.02}$ & $0.29{\scriptstyle\pm0.00}$ & $0.66{\scriptstyle\pm0.01}$ \\
MT-PPO & $0.40{\scriptstyle\pm0.00}$ & $0.80{\scriptstyle\pm0.01}$ & $0.56{\scriptstyle\pm0.01}$ & $0.41{\scriptstyle\pm0.01}$ & $0.26{\scriptstyle\pm0.00}$ & $0.55{\scriptstyle\pm0.01}$ \\
ArCHer & $0.41{\scriptstyle\pm0.01}$ & $0.67{\scriptstyle\pm0.03}$ & $0.57{\scriptstyle\pm0.01}$ & $\boldsymbol{0.56{\scriptstyle\pm0.02}}$ & $0.20{\scriptstyle\pm0.00}$ & $0.34{\scriptstyle\pm0.01}$ \\
ArCHer-LC & $0.37{\scriptstyle\pm0.01}$ & $0.61{\scriptstyle\pm0.02}$ & $0.54{\scriptstyle\pm0.02}$ & $0.43{\scriptstyle\pm0.03}$ & $0.18{\scriptstyle\pm0.02}$ & $0.31{\scriptstyle\pm0.02}$ \\
ST-GRPO & $0.55{\scriptstyle\pm0.00}$ & $0.93{\scriptstyle\pm0.03}$ & $0.60{\scriptstyle\pm0.03}$ & $0.46{\scriptstyle\pm0.01}$ & $0.29{\scriptstyle\pm0.04}$ & $0.61{\scriptstyle\pm0.03}$ \\
\midrule
IterPO & $\boldsymbol{0.56{\scriptstyle\pm0.00}}$ & $\boldsymbol{0.99{\scriptstyle\pm0.01}}$ & $\boldsymbol{0.64{\scriptstyle\pm0.02}}$ & $0.48{\scriptstyle\pm0.00}$ & $\boldsymbol{0.36{\scriptstyle\pm0.03}}$ & $\boldsymbol{0.71{\scriptstyle\pm0.01}}$ \\
\bottomrule
\end{tabular}
\end{table*}

\begin{table*}[ht]
\centering
\caption{Qwen7b in-training expected discounted return $J(\pi)$ at the final round (Round~4). ``Base'' reports the initial instruction-tuned policy evaluated before training; trained methods report Round~4. Means $\pm 1$ SE across 3 seeds. The held-out counterpart is Table~\ref{tab:summary_test_r4}.}
\label{tab:summary_r4}
\small
\setlength{\tabcolsep}{4pt}
\begin{tabular}{lcccccc}
\toprule
Method & Craigslist & Ultimatum & DealOrNoDeal & CaSiNo & Job Interview & Alliance \\
\midrule
Base & $0.38$ & $0.77$ & $0.55$ & $0.39$ & $0.26$ & $0.55$ \\
\midrule
MT-SFT & $0.42{\scriptstyle\pm0.00}$ & $0.79{\scriptstyle\pm0.01}$ & $0.26{\scriptstyle\pm0.02}$ & $0.37{\scriptstyle\pm0.01}$ & $0.25{\scriptstyle\pm0.01}$ & $0.54{\scriptstyle\pm0.01}$ \\
MT-KTO & $0.49{\scriptstyle\pm0.02}$ & $0.86{\scriptstyle\pm0.01}$ & $0.57{\scriptstyle\pm0.02}$ & $0.53{\scriptstyle\pm0.02}$ & $0.29{\scriptstyle\pm0.00}$ & $0.65{\scriptstyle\pm0.01}$ \\
MT-PPO & $0.40{\scriptstyle\pm0.00}$ & $0.80{\scriptstyle\pm0.01}$ & $0.55{\scriptstyle\pm0.02}$ & $0.39{\scriptstyle\pm0.01}$ & $0.26{\scriptstyle\pm0.00}$ & $0.53{\scriptstyle\pm0.02}$ \\
ArCHer & $0.41{\scriptstyle\pm0.00}$ & $0.67{\scriptstyle\pm0.03}$ & $0.57{\scriptstyle\pm0.01}$ & $\boldsymbol{0.56{\scriptstyle\pm0.02}}$ & $0.19{\scriptstyle\pm0.01}$ & $0.30{\scriptstyle\pm0.01}$ \\
ArCHer-LC & $0.37{\scriptstyle\pm0.01}$ & $0.61{\scriptstyle\pm0.02}$ & $0.50{\scriptstyle\pm0.01}$ & $0.41{\scriptstyle\pm0.06}$ & $0.17{\scriptstyle\pm0.01}$ & $0.31{\scriptstyle\pm0.02}$ \\
ST-GRPO & $0.55{\scriptstyle\pm0.00}$ & $0.93{\scriptstyle\pm0.03}$ & $0.39{\scriptstyle\pm0.14}$ & $0.46{\scriptstyle\pm0.05}$ & $0.29{\scriptstyle\pm0.04}$ & $0.61{\scriptstyle\pm0.03}$ \\
\midrule
IterPO & $\boldsymbol{0.56{\scriptstyle\pm0.01}}$ & $\boldsymbol{0.98{\scriptstyle\pm0.00}}$ & $\boldsymbol{0.64{\scriptstyle\pm0.02}}$ & $0.48{\scriptstyle\pm0.00}$ & $\boldsymbol{0.36{\scriptstyle\pm0.03}}$ & $\boldsymbol{0.66{\scriptstyle\pm0.02}}$ \\
\bottomrule
\end{tabular}
\end{table*}

\section{Held-Out Kernels}
\label{appendix:held_out_test}

\input{appendix_held_out_test}

Table~\ref{tab:summary_test_r4} reports held-out return at Round~4.\par

\begin{table*}[ht]
\centering
\caption{Qwen7b expected discounted return $J(\pi)$ on \textbf{held-out} kernels at the \textbf{final round} (Round~4). ``Base'' reports the initial instruction-tuned policy; trained methods report their Round-4 checkpoints. Means $\pm 1$ SE across 3 seeds.}
\label{tab:summary_test_r4}
\small
\setlength{\tabcolsep}{4pt}
\begin{tabular}{lcccccc}
\toprule
Method & Craigslist & Ultimatum & DealOrNoDeal & CaSiNo & Job Interview & Alliance \\
\midrule
Base & $0.36$ & $0.80$ & $0.52$ & $0.38$ & $0.25$ & $0.48$ \\
\midrule
MT-SFT & $0.41{\scriptstyle\pm0.01}$ & $0.80{\scriptstyle\pm0.01}$ & $0.24{\scriptstyle\pm0.03}$ & $0.40{\scriptstyle\pm0.02}$ & $0.25{\scriptstyle\pm0.01}$ & $0.49{\scriptstyle\pm0.00}$ \\
MT-KTO & $0.48{\scriptstyle\pm0.01}$ & $0.86{\scriptstyle\pm0.01}$ & $0.55{\scriptstyle\pm0.04}$ & $\boldsymbol{0.54{\scriptstyle\pm0.01}}$ & $0.29{\scriptstyle\pm0.01}$ & $\boldsymbol{0.60{\scriptstyle\pm0.01}}$ \\
MT-PPO & $0.38{\scriptstyle\pm0.01}$ & $0.80{\scriptstyle\pm0.02}$ & $0.56{\scriptstyle\pm0.02}$ & $0.41{\scriptstyle\pm0.02}$ & $0.25{\scriptstyle\pm0.01}$ & $0.51{\scriptstyle\pm0.02}$ \\
ArCHer & $0.44{\scriptstyle\pm0.02}$ & $0.91{\scriptstyle\pm0.03}$ & $0.60{\scriptstyle\pm0.03}$ & $0.53{\scriptstyle\pm0.03}$ & $0.28{\scriptstyle\pm0.01}$ & $0.53{\scriptstyle\pm0.02}$ \\
ArCHer-LC & $0.38{\scriptstyle\pm0.01}$ & $0.86{\scriptstyle\pm0.02}$ & $0.58{\scriptstyle\pm0.03}$ & $0.40{\scriptstyle\pm0.03}$ & $0.25{\scriptstyle\pm0.02}$ & $0.49{\scriptstyle\pm0.06}$ \\
ST-GRPO & $0.54{\scriptstyle\pm0.00}$ & $0.94{\scriptstyle\pm0.05}$ & $0.36{\scriptstyle\pm0.14}$ & $0.45{\scriptstyle\pm0.03}$ & $0.28{\scriptstyle\pm0.04}$ & $0.55{\scriptstyle\pm0.02}$ \\
\midrule
IterPO & $\boldsymbol{0.55{\scriptstyle\pm0.01}}$ & $\boldsymbol{0.99{\scriptstyle\pm0.01}}$ & $\boldsymbol{0.70{\scriptstyle\pm0.01}}$ & $0.45{\scriptstyle\pm0.03}$ & $\boldsymbol{0.35{\scriptstyle\pm0.03}}$ & $0.51{\scriptstyle\pm0.04}$ \\
\bottomrule
\end{tabular}
\end{table*}


\section{Additional Backbone: Llama3b}
\label{appendix:llama3b}

\input{appendix_llama3b}

\clearpage

\section{Systematic Behavioral Analysis}
\label{appendix:behavioral}

\input{appendix_behavioral_analysis}

%% file: appendix_craigslist.tex

\subsubsection{Conversation Generation}

Algorithm~\ref{alg:craigslist_turn} summarizes the conversation generation.

\begin{algorithm}[t]
\caption{Craigslist Negotiation: Conversation Generation}
\label{alg:craigslist_turn}
\begin{algorithmic}[1]
\STATE \textbf{Input:} Scenario $k$ with listing price $p_{\text{list}}$ and buyer target, vendor policy $\pi$
\STATE $\texttt{turns} \leftarrow [\,]$
\STATE $m_0 \leftarrow \texttt{VendorOpeningMessage}(k)$ \hfill \textit{// ``I'm selling X for \$Y...''}
\STATE Append $(\textnormal{vendor}, m_0)$ to $\texttt{turns}$
\FOR{$t = 1$ to $\texttt{max\_turns}$}
    \STATE $\texttt{role} \leftarrow \texttt{customer}$ if $t$ is odd else $\texttt{vendor}$
    \IF{$\texttt{role} = \texttt{vendor}$}
        \STATE Activate LoRA adapter
    \ELSE
        \STATE Activate base model
    \ENDIF
    \STATE $\texttt{prompt} \leftarrow \texttt{BuildPrompt}(k, \texttt{turns}, \texttt{role})$
    \STATE $m_t \leftarrow \texttt{LLM}(\texttt{prompt})$
    \STATE Append $(\texttt{role}, m_t)$ to $\texttt{turns}$
    \IF{$m_t$ contains \texttt{\textbackslash boxed\{AGREED: \$X\}}}
        \STATE \textbf{return} $(\texttt{turns}, r=X / p_{\text{list}}, \texttt{done}=\texttt{True})$
    \ELSIF{$m_t$ contains \texttt{\textbackslash boxed\{NO DEAL\}}}
        \STATE \textbf{return} $(\texttt{turns}, r=0, \texttt{done}=\texttt{True})$
    \ENDIF
\ENDFOR
\STATE \textbf{return} $(\texttt{turns}, r=0, \texttt{done}=\texttt{True})$ \hfill \textit{// Max turns reached}
\end{algorithmic}
\end{algorithm}

\subsubsection{Agent Prompts}

Both agents receive short system prompts that specify their roles and private pricing information. The vendor sees only the item description and listing price and receives no explicit strategy guidance. The customer also sees a target price derived from the original human negotiation. Both prompts require the agent to mark an agreement with \texttt{\textbackslash boxed\{AGREED: \$X\}} or end the negotiation with \texttt{\textbackslash boxed\{NO DEAL\}}.\par

\vspace{5pt}

\begin{promptbox}[title=Vendor System Prompt]
\small\ttfamily
You are selling: [item description]\\[6pt]
Your listing price is \$[listing price]. Negotiate to get the best possible price.\\[6pt]
\textbf{RULES:}\\
- When you reach an agreement, respond with exactly: \textbackslash boxed\{AGREED: \$<price>\}\\
- If no agreement can be reached, respond with: \textbackslash boxed\{NO DEAL\}
\end{promptbox}

\vspace{5pt}

\begin{promptbox}[title=Customer System Prompt]
\small\ttfamily
You are buying: [item description]\\[6pt]
The listing price is \$[listing price]. Your target price is \$[buyer target].\\[6pt]
\textbf{RULES:}\\
- Negotiate to get as close to your target price of \$[buyer target] as possible\\
- When you reach an agreement, respond with exactly: \textbackslash boxed\{AGREED: \$<price>\}\\
- If the price is too high, respond with: \textbackslash boxed\{NO DEAL\}
\end{promptbox}

The vendor starts each negotiation with a scripted opening:

\begin{promptbox}[title=Vendor Opening Message]
\small\ttfamily
Hello! I'm selling [item description].\\[6pt]
My asking price is \$[listing price]. Are you interested?
\end{promptbox}

\subsubsection{Data Source and Economic Parameters}

The CraigslistBargain dataset \citep{he2018decoupling} provides real listing information (item titles, descriptions, listing prices) and buyer target prices from the original human negotiations across six product categories: housing, furniture, cars, bikes, phones, and electronics. Each negotiation scenario uses two parameters from the dataset:

\begin{itemize}
    \item \textbf{Listing price} ($p_{\text{list}}$): The vendor's asking price, taken directly from the Craigslist listing.
    \item \textbf{Buyer target} ($p_{\text{target}}$): The buyer's target price, from the original human negotiation data.
\end{itemize}

The vendor sees only the listing price; the customer sees both the listing price and their target. No additional economic parameters (reservation prices, cost bases, budgets) are provided to either agent. The vendor must discover its negotiation strategy entirely through reinforcement learning.

\subsubsection{Reward Structure}

We use a \emph{price ratio} reward that measures the agreed price relative to the listing price:
\[
R(s_\tau, a_\tau, x_{\tau+1}) = \begin{cases}
\dfrac{p}{p_{\text{list}}} & \text{if agreement at price } p \\[10pt]
0 & \text{if no agreement}
\end{cases}
\]

The reward assigns zero to failed negotiations and scales successful ones by the agreed price, so it rewards both agreement and higher sale prices. For example, a sale at the listing price yields $1.0$; a \$32 sale on a \$40 listing yields $32/40=0.8$.\par

%% file: appendix_ultimatum.tex

\subsubsection{Agent Prompts}

In the Ultimatum Game, a Proposer ($\pi$, the trained agent) divides a resource pool with a Responder ($\pi_{\text{env}}$, the static environment agent), who may accept, reject, or counter-propose. Parallel system prompts specify each agent's role, the pool size, the round limit, and the terminal markers. We show the Proposer prompt below. The Responder prompt instead assigns the agent to Player~2 and retains the convention that \texttt{\$X} denotes Player~1's share.\par

\vspace{5pt}

\begin{promptbox}[title=Proposer System Prompt ($\pi$)]
\small\ttfamily
You are Player 1 (the Proposer) in a resource-splitting game.\\[6pt]
There is a pool of \$[pool\_size] to divide between you and Player 2. You start with the entire pool; Player 2 starts with nothing. You will negotiate over [num\_rounds] total rounds (alternating turns).\\[6pt]
IMPORTANT: If no agreement is reached by the end of [num\_rounds] rounds, BOTH players get NOTHING. Rejection destroys the entire pool.\\[6pt]
On your final turn, you must either Accept or Reject (no counter-proposals).\\[6pt]
\textbf{RULES:}\\
- Propose splits, counter-propose, or accept the other player's proposal\\
- When you reach an agreement, respond with exactly: \textbackslash boxed\{AGREED: split=\$X\} where X is YOUR share\\
- To reject and end the game (both get zero), respond with: \textbackslash boxed\{NO DEAL\}
\end{promptbox}

\subsubsection{Opening Message}

The Proposer starts each negotiation with a scripted opening message:

\begin{promptbox}[title=Proposer Opening Message]
\small\ttfamily
Hello! We need to split \$[pool\_size] between us. I hold the pool and will make the first proposal. If we can't agree, we both get nothing. What do you think would be a fair split?
\end{promptbox}

\subsubsection{Data Source}

We generate scenarios from every combination of five pool sizes and three round limits:\par
\begin{itemize}
    \item \textbf{Pool size} ($p$): drawn from $\{100, 500, 1000, 5000, 10000\}$.
    \item \textbf{Number of rounds}: drawn from $\{4, 6, 8\}$.
\end{itemize}
The Cartesian product yields 15 configurations. When a training round requires more than 15 scenarios, we cycle through them. This construction follows the Ultimatum Game formulation in NegotiationArena~\citep{bianchi2024negotiationarena}.\par

\subsubsection{Reward Structure}

The reward measures the Proposer's agreed share relative to a baseline of an equal split:
\[
R(s_\tau, a_\tau, x_{\tau+1}) = \begin{cases}
\dfrac{X}{p / 2} & \text{if agreement with Proposer share } X \\[10pt]
0 & \text{if no agreement (}\texttt{\textbackslash boxed\{NO DEAL\}}\text{)}
\end{cases}
\]
where $p$ is the pool size. An equal split ($X = p/2$) yields a reward of $1.0$. Taking the entire pool ($X = p$) yields a reward of $2.0$. A no-deal outcome, where both players receive nothing, yields a reward of $0$.

%% file: appendix_deal_or_no_deal.tex

\subsubsection{Agent Prompts}

Both agents receive parallel system prompts that describe the item pool, private valuations, and bilateral-acceptance protocol. The trained agent $\pi$ observes only its own private values, while the environment agent $\pi_{\text{env}}$ observes a different set of private values. Neither agent observes the other's valuations.\par

\vspace{5pt}

\begin{promptbox}[title=System Prompt (both agents)]
\small\ttfamily
You are negotiating to divide items with another person. You each have different private values for the items --- the other person's values are unknown to you.\\[6pt]
The items are: book=(count:X value:Y) hat=(count:X value:Y) ball=(count:X value:Y)\\[6pt]
Your goal is to maximize your total points. You get points only for items YOU receive. All items must be allocated between the two of you.\\[6pt]
\textbf{PROCESS:}\\
Step 1 --- Discuss preferences in plain text. Share which items you value most and least, and ask the other person about theirs. Do not write the AGREED box yet.\\[4pt]
Step 2 --- Once you both understand each other's preferences, propose an allocation by writing the AGREED box. Both sides may make counter-proposals; iterate until you converge.\\[6pt]
\textbf{PROTOCOL --- how deals close:}\\[4pt]
1. PROPOSE an allocation by writing:\\
\quad\textbackslash boxed\{AGREED: MINE: book=$n_b$ hat=$n_h$ ball=$n_l$; THEIRS: book=$(B{-}n_b)$ hat=$(H{-}n_h)$ ball=$(L{-}n_l)$\}\\
where MINE is the share YOU receive and THEIRS is the share the OTHER person receives. Use actual integers in place of the $n_*$ placeholders. REQUIRED: include ALL THREE items in BOTH MINE and THEIRS, in book/hat/ball order, even when a count is 0. The two sides must partition all items exactly.\\[4pt]
2. To ACCEPT the other person's most recent PROPOSE, write exactly:\\
\quad\textbackslash boxed\{ACCEPT\}\\
This seals the deal at the partition THEY proposed.\\[4pt]
3. To REJECT the standing proposal without closing the deal, write:\\
\quad\textbackslash boxed\{REJECT\}\\
Then continue negotiating in plain text or with a new PROPOSE.\\[4pt]
4. A deal closes ONLY when one party writes a PROPOSE box and the OTHER party writes \textbackslash boxed\{ACCEPT\}. Plain text like ``I agree'' or silence does NOT close a deal. Writing a new PROPOSE supersedes any standing proposal.\\[4pt]
5. If no agreement can be reached at all, write: \textbackslash boxed\{NO DEAL\}\\[4pt]
6. If you do not feel the standing proposal is the best outcome for you, negotiate further in plain text until you are able to find a better outcome.
\end{promptbox}

\paragraph{Per-turn private-values reminder.}
Before each generation, we prepend a one-line reminder of the active agent's private values and the item pool to the most recent user message (e.g., \texttt{[REMINDER --- your private values: book=1 hat=4 ball=2. Pool: book=4 hat=1 ball=1. Maximize MINE$\cdot$values.]}). The reminder keeps these values in the immediate context because base models otherwise drift from their assigned valuations over long dialogues.\par

\paragraph{Validation.}
The environment validates every AGREED proposal box before treating it as the standing proposal. All three item types (book, hat, ball) must appear in both MINE and THEIRS, every count must be a non-negative integer, and MINE$_i$ + THEIRS$_i$ must equal the pool count for each item $i$. An invalid box remains ordinary dialogue and cannot be the target of an ACCEPT. The bilateral rule requires an explicit ACCEPT or verbatim echo from the other party, so both agents must commit to the same partition.\par

\subsubsection{Opening Message}

Agent~1 ($\pi$) starts each negotiation with a scripted opening message that states the item counts without revealing any value information.

\begin{promptbox}[title=Opening Message]
\small\ttfamily
Hello! We need to divide some items between us: X books, Y hats, Z balls. Let's discuss how to split them fairly. What items are you most interested in?
\end{promptbox}

\subsubsection{Data Source}

Scenarios come from the Deal or No Deal dataset of \citet{lewis2017deal}. We load the \texttt{dialogues} configuration of \texttt{mikelewis0/deal\_or\_no\_dialog} from Hugging Face. The dataset contains approximately 5{,}000 negotiation dialogues collected through Amazon Mechanical Turk. Each scenario specifies the counts of books, hats, and balls and assigns a private valuation vector to each agent. The construction sets each agent's total value across the item pool to 10, so the agents have equal maximum attainable utility but different preferences over the items.\par

\subsubsection{Reward Structure}

When a deal closes, we recover the trained agent's allocation from the accepted proposal. We use MINE when the trained agent proposed the allocation and THEIRS when the partner proposed it. The reward is\par
\[
R(s_\tau, a_\tau, x_{\tau+1}) = \begin{cases}
\displaystyle\frac{\sum_{i \in \{\text{book}, \text{hat}, \text{ball}\}} a_i \cdot v_i}{U_{\max} / 2} & \text{if agreement} \\[10pt]
0 & \text{if no deal}
\end{cases}
\]
where $a_i$ is the number of units of item $i$ allocated to the trained agent, $v_i$ is its private value for that item, and $U_{\max}=\sum_i c_i v_i$ is the utility obtained from receiving the entire pool. Dividing by $U_{\max}/2$ assigns a reward of $1.0$ to a split worth half the maximum utility and $2.0$ to receiving every item. In the standard Deal or No Deal dataset, $U_{\max}=10$ for every agent and scenario, so the implementation uses a denominator of $5$.\par

An invalid PROPOSE box remains ordinary dialogue and cannot be accepted. A conversation that ends without a valid PROPOSE/ACCEPT pair receives $R=0$.\par

%% file: appendix_casino.tex

\subsubsection{Agent Prompts}
\label{appendix:casino:prompts}

\looseness-1 The trained agent ($\pi$, Camper~1) and static environment agent ($\pi_{\text{env}}$, Camper~2) receive parallel system prompts containing the bilateral-acceptance protocol and different private priorities and participant-written reasons for Food, Water, and Firewood.\par

\vspace{5pt}

\begin{promptbox}[title=System Prompt (both campers)]
\small\ttfamily
Imagine that you are on a camping trip. Apart from some basic supplies, you can collect additional food packages, water bottles and firewood to make your trip better. Since these are limited, you will have to split them with your campsite neighbor.\\[6pt]
There are 3 packages each of Food, Water, and Firewood (9 total) to divide.\\[6pt]
Your priorities:\\
- [Issue] is [High/Medium/Low] priority ([reason])\\
- [Issue] is [High/Medium/Low] priority ([reason])\\
- [Issue] is [High/Medium/Low] priority ([reason])\\[6pt]
Points per package: High priority=5, Medium=4, Low=3. Maximize your points. Try hard to get as many items as you can!\\[6pt]
\textbf{PROCESS:}\\
Step 1 --- Discuss preferences in plain text. Share which items you prioritize highest and lowest, and ask the other person about theirs.\\[4pt]
Step 2 --- Once you both understand each other's preferences, propose an allocation by writing the AGREED box. Both sides may make counter-proposals; iterate until you converge.\\[6pt]
\textbf{PROTOCOL --- how deals close:}\\[4pt]
1. PROPOSE an allocation by writing:\\
\quad\textbackslash boxed\{AGREED: MINE: Food=$n_f$ Water=$n_w$ Firewood=$n_{fw}$; THEIRS: Food=$(3{-}n_f)$ Water=$(3{-}n_w)$ Firewood=$(3{-}n_{fw})$\}\\
where MINE is your share and THEIRS is the other person's. Use integers in $\{0,1,2,3\}$; the two sides must partition all 9 packages exactly.\\[4pt]
2. To ACCEPT the other person's most recent PROPOSE, write exactly:\\
\quad\textbackslash boxed\{ACCEPT\}\\
3. To REJECT the standing proposal without closing the deal, write:\\
\quad\textbackslash boxed\{REJECT\}\\
4. A deal closes ONLY when one party writes a PROPOSE box and the OTHER party writes \textbackslash boxed\{ACCEPT\}. Plain text like ``I agree'' or silence does NOT close a deal. Writing a new PROPOSE supersedes any standing proposal.\\[4pt]
5. If no agreement can be reached at all, write: \textbackslash boxed\{NO DEAL\}\\[4pt]
6. If you do not feel the standing proposal is the best outcome for you, negotiate further in plain text until you are able to find a better outcome.
\end{promptbox}

\looseness-1 Each scenario assigns every camper one High-, one Medium-, and one Low-priority resource. We fill the bracketed fields with these assignments and the participant-written reasons recorded verbatim in the dataset.\par

\paragraph{Per-turn private-priority reminder.}
Before each generation, we prepend a one-line reminder of the active agent's private priorities, their point values, and the resource pool to the most recent user message (e.g., \texttt{[REMINDER --- your private priorities: Food=High(5pts) Water=Medium(4pts) Firewood=Low(3pts). Pool: Food=3 Water=3 Firewood=3. Maximize MINE$\cdot$points.]}). The reminder keeps these priorities in the immediate context over long dialogues.\par

\paragraph{Validation.}
\looseness-1 The environment validates every AGREED proposal box before treating it as the standing proposal. Food, Water, and Firewood must appear in both MINE and THEIRS, every count must be a non-negative integer, and MINE$_i$ + THEIRS$_i$ must equal 3 for each resource $i$. An invalid box remains ordinary dialogue and cannot be accepted. A deal closes only after the other camper writes an explicit ACCEPT or repeats the proposal verbatim, so both campers commit to the same allocation.\par

\subsubsection{Opening Message}
\label{appendix:casino:opening}

Camper~1 ($\pi$) starts each negotiation with a scripted opening message:

\begin{promptbox}[title=Camper 1 Opening Message]
\small\ttfamily
Hi there! Looks like we need to split up some camping supplies. I have some preferences about what I'd like. How about you? What are you looking for most?
\end{promptbox}

\subsubsection{Data Source}
\label{appendix:casino:data}

Scenarios come from the CaSiNo dataset \citep{chawla2021casino}, which contains 1,030 campsite negotiation dialogues collected through Amazon Mechanical Turk. Each record provides both campers' High, Medium, and Low priority assignments over Food, Water, and Firewood, along with participant-written reasons (e.g., ``Water is High priority because I get dehydrated easily'').\par

\looseness-1 We load \texttt{kchawla123/casino} from Hugging Face and partition its single split 85/15 into in-training and held-out partitions. Each scenario uses both campers' priority assignments and associated reasons.\par

\subsubsection{Reward Structure}
\label{appendix:casino:reward}

When a deal closes, we recover the trained agent's allocation from the accepted proposal. We use MINE when the trained agent proposed the allocation and THEIRS when the partner proposed it. The trained agent receives 5 points per package of its High-priority resource, 4 per package of its Medium-priority resource, and 3 per package of its Low-priority resource.\par

The reward is
\[
R(s_\tau, a_\tau, x_{\tau+1}) = \begin{cases}
\dfrac{\displaystyle\sum_{i \in \{\text{Food}, \text{Water}, \text{Firewood}\}} n_i \cdot w_i}{U_{\max} / 2} & \text{if agreement} \\[14pt]
0 & \text{if no deal}
\end{cases}
\]
where $n_i \in \{0,1,2,3\}$ is the number of packages of resource $i$ allocated to the trained agent and $w_i \in \{5,4,3\}$ is the corresponding priority weight. Because every camper has one High-, one Medium-, and one Low-priority resource, receiving all nine packages gives $U_{\max}=3(5+4+3)=36$ in every scenario. The implementation therefore divides by $U_{\max}/2=18$, assigning a reward of $1.0$ to a split worth half the maximum utility and $2.0$ to receiving every package.\par

An invalid PROPOSE box remains ordinary dialogue and cannot be accepted. A conversation that ends without a valid PROPOSE/ACCEPT pair receives $R=0$.\par

%% file: appendix_job_interview.tex

\subsubsection{Agent Prompts}

The Job Interview environment pairs a trained Recruiter policy $\pi$ with a static Worker policy $\pi_{\text{env}}$. The agents negotiate Salary, Weekly Holiday, Workplace, Company, and Position. The Recruiter prefers lower Salary and Weekly Holiday values, while the Worker prefers higher values; each agent separately values the three discrete issues. Both agents receive parallel system prompts that specify their role, the five issues, and a procedurally generated utility function. We show the Recruiter prompt below and describe the Worker-specific changes afterward.\par

\vspace{5pt}

\begin{promptbox}[title=Recruiter System Prompt ($\pi$)]
\small\ttfamily
You are a recruiter negotiating a job offer with a candidate. You need to agree on 5 issues: Salary (20-51, in thousands), Weekly Holiday (2-7 days), Workplace (Tokyo/Seoul/Beijing/Sydney), Company (Google/Facebook/Apple/Amazon), and Position (Engineer/Manager/Designer/Sales).\\[6pt]
Your preferences (importance weights and option values):\\[6pt]
- Salary (range 20-51, you prefer LOWER): importance = [weight]\\
- WeeklyHoliday (range 2-7, you prefer LOWER): importance = [weight]\\
- Workplace (importance = [weight]): Tokyo=[val], Seoul=[val], Beijing=[val], Sydney=[val]\\
- Company (importance = [weight]): Google=[val], Facebook=[val], Apple=[val], Amazon=[val]\\
- Position (importance = [weight], value depends on Company chosen):\\
\hspace*{1em}If Company=Google: Engineer=[val], Manager=[val], Designer=[val], Sales=[val]\\
\hspace*{1em}If Company=Facebook: Engineer=[val], Manager=[val], Designer=[val], Sales=[val]\\
\hspace*{1em}If Company=Apple: Engineer=[val], Manager=[val], Designer=[val], Sales=[val]\\
\hspace*{1em}If Company=Amazon: Engineer=[val], Manager=[val], Designer=[val], Sales=[val]\\[6pt]
Your goal is to maximize your total score. The candidate has different preferences. Try to find a deal that works for both of you.\\[6pt]
\textbf{RULES:}\\
- When you reach an agreement, respond with exactly: \textbackslash boxed\{AGREED: Salary=X Holiday=Y Workplace=Z Company=W Position=P\}\\
- If no agreement can be reached, respond with: \textbackslash boxed\{NO DEAL\}
\end{promptbox}

The Worker prompt changes the role from recruiter to job candidate and reverses the direction of preference for the integer issues: the Worker prefers higher Salary and Weekly Holiday values, while the Recruiter prefers lower values. We sample each party's values for Workplace, Company, and Position independently, so their discrete preferences may agree or conflict.\par

\subsubsection{Opening Message}

The Recruiter starts each negotiation with a scripted opening:

\begin{promptbox}[title=Recruiter Opening Message]
\small\ttfamily
Hello! I'm the recruiter and I'd like to discuss the terms of your offer. We need to agree on salary, weekly holiday, workplace location, company, and position. What matters most to you?
\end{promptbox}

\subsubsection{Data Source}

We adapt the scenario structure from the multi-issue negotiation framework of \citet{yamaguchi2021eacl}. For each scenario, we independently sample the Recruiter's and Worker's utility functions. The resulting pair of utility functions defines the scenario's Pareto frontier.\par

The agents negotiate two integer-valued issues and three discrete issues:\par
\begin{itemize}
    \item \textbf{Salary} (integer, 20-51): salary in thousands. The Recruiter prefers lower values, while the Worker prefers higher values.
    \item \textbf{Weekly Holiday} (integer, 2-7): days off per week. The Recruiter prefers lower values, while the Worker prefers higher values.
    \item \textbf{Workplace} (discrete): Tokyo, Seoul, Beijing, or Sydney.
    \item \textbf{Company} (discrete): Google, Facebook, Apple, or Amazon.
    \item \textbf{Position} (discrete): Engineer, Manager, Designer, or Sales. The value of a Position depends on the selected Company, introducing a cross-issue dependency.
\end{itemize}

We generate each party's utility function independently. We draw five issue-importance weights uniformly, normalize them, rescale them to $[0.1, 0.6]$, and renormalize them to sum to one. For Workplace and Company, we draw option values and min-max normalize each issue's values to $[0,1]$. For Position, we draw and normalize a separate set of values for each Company, yielding a $4 \times 4$ Company-by-Position table.\par
This procedure generates an unlimited supply of unique scenarios.

\subsubsection{Reward Structure}

For a valid agreement, the Recruiter's reward is the weighted sum of its normalized per-issue scores:\par
\[
R(s_\tau, a_\tau, x_{\tau+1}) = \begin{cases}
\displaystyle\sum_{i \in \mathcal{I}} w_i \cdot \text{score}_i & \text{if the agreement is valid} \\[10pt]
0 & \text{otherwise}
\end{cases}
\]
where $\mathcal{I} = \{\text{Salary}, \text{WeeklyHoliday}, \text{Workplace}, \text{Company}, \text{Position}\}$ and $w_i$ is the importance weight for issue $i$ (with $\sum_i w_i = 1$). The per-issue scores are computed as follows:

\begin{itemize}
    \item \textbf{Integer issues} (Salary, WeeklyHoliday): the recruiter prefers lower values, so $\text{score}_i = (\text{max} - \text{chosen}) / (\text{max} - \text{min})$.
    \item \textbf{Discrete issues} (Workplace, Company): $\text{score}_i$ equals the option value from the recruiter's utility function (already in $[0, 1]$).
    \item \textbf{Dependent issue} (Position): $\text{score}_i$ equals the position-company bias value for the chosen (Company, Position) pair.
\end{itemize}

Because the issue weights sum to one and every per-issue score lies in $[0,1]$, the reward also lies in $[0,1]$. A no-deal outcome, signaled by \texttt{\textbackslash boxed\{NO DEAL\}}, receives a reward of zero. Invalid agreements also receive zero; an agreement is invalid if Salary falls outside $[20,51]$, Weekly Holiday falls outside $[2,7]$, or any discrete choice is unrecognized.\par

%% file: appendix_alliance.tex

\subsubsection{Agent Prompts}

\looseness-1 The Alliance environment pairs a trained Stakeholder~1 policy $\pi$ with a static Stakeholder~2 policy $\pi_{\text{env}}$. Each agent's system prompt lists the five issues, the available options, its private option scores, and a minimum acceptable score that represents its best alternative to a negotiated agreement (BATNA). Each agent sees only its own scores and threshold. We show Stakeholder~1's prompt below and describe Stakeholder~2's prompt afterward.\par

\vspace{5pt}

\begin{promptbox}[title=Stakeholder 1 ($\pi$) System Prompt]
\small\ttfamily
You are a stakeholder negotiating a project agreement with another party. You each have private preferences over the options for each issue.\\[6pt]
The issues and options are:\\
Issue A (Infrastructure): A1=Water-based facilities, A2=Land-based facilities\\
Issue B (Environment): B1=Minimal impact, B2=Moderate restoration, B3=Full restoration\\
Issue C (Employment): C1=Unlimited hiring, C2=Preference to locals, C3=Partial union, C4=Full union\\
Issue D (Funding): D1=Full federal loan, D2=Partial loan, D3=Minimal loan, D4=No federal loan\\
Issue E (Compensation): E1=Maximum compensation, E2=High compensation, E3=Moderate compensation, E4=Low compensation, E5=No compensation\\[6pt]
Your private scores for each option:\\[6pt]
Issue A (Infrastructure): A1 (Water-based facilities): X pts, A2 (Land-based facilities): Y pts\\
Issue B (Environment): B1 (Minimal impact): X pts, B2 (Moderate restoration): Y pts, B3 (Full restoration): Z pts\\
\ldots\\[6pt]
Your minimum acceptable score (BATNA threshold): [threshold]\\
You should REJECT any deal that gives you fewer than this many points.\\[6pt]
Your goal is to maximize your score while ensuring it stays above your threshold. The other party has different (likely opposing) preferences.\\[6pt]
\textbf{RULES:}\\
- When you reach an agreement, respond with exactly: \textbackslash boxed\{AGREED: A=1 B=2 C=3 D=1 E=4\} using issue letters and option numbers (1-indexed)\\
- If no agreement can be reached, respond with: \textbackslash boxed\{NO DEAL\}
\end{promptbox}

\vspace{5pt}

Stakeholder~2 receives the same prompt with its own private scores and BATNA threshold.\par

\subsubsection{Opening Message}

Stakeholder~1 ($\pi$) starts each negotiation with a scripted opening message:

\begin{promptbox}[title=Stakeholder 1 Opening Message]
\small\ttfamily
Hello! We need to agree on the terms of this project. There are several issues to negotiate and I'm sure we can find something that works for both of us. Which issues are most important to you?
\end{promptbox}

\subsubsection{Data Source}

We adapt the Alliance environment from the LLM-Deliberation framework of \citet{abdelnabi2024llm}. We retain its five issue labels and option counts: Infrastructure, Environment, Employment, Funding, and Compensation have 2, 3, 4, 4, and 5 options, respectively. We generate the score matrices and BATNA thresholds using the procedure below.\par

\looseness-1 For each party, we randomly allocate a 100-point budget across the five issues and draw scores for the options within each issue. Both parties assign the highest score on each issue to its first option. The deal that selects the first option for all five issues therefore maximizes both parties' scores. Each agent sees only its own score matrix and is told that the other party's preferences are likely to oppose its own.\par

We sample each agent's BATNA threshold from a difficulty-dependent fraction of its feasible score range:\par
\begin{itemize}
    \item \textbf{Easy}: $[0.2, 0.35]$.
    \item \textbf{Medium}: $[0.35, 0.55]$.
    \item \textbf{Hard}: $[0.55, 0.70]$.
\end{itemize}

After sampling both thresholds, we check whether at least one deal meets both. If none does, we lower the thresholds until a feasible deal exists.\par
The generator produces an unlimited number of unique scenarios.

\subsubsection{Reward Structure}

Let
\[
U = \sum_{i \in \{A,B,C,D,E\}} \text{score}_i[\text{option}_i]
\]
denote Stakeholder~1's utility for a proposed deal, and let $b$ denote its BATNA threshold. The reward is
\[
R(s_\tau, a_\tau, x_{\tau+1}) =
\begin{cases}
U/50 & \text{if the agreement is valid and } U \geq b, \\[6pt]
0 & \text{otherwise}.
\end{cases}
\]
The maximum possible utility is 100, so dividing $U$ by 50 assigns a reward of 1 to a 50-point deal and allows rewards up to 2. For example, an agreement with $U=65$ and $b=40$ receives a reward of $65/50=1.3$. An agreement with $U<b$ receives zero because it is worse than the agent's outside option.\par

%% file: appendix_hyperparameters.tex
We use the same hyperparameters across all six environments and run every configuration with three random seeds. The full settings for each method appear below. All methods train on two 40 GB A100 GPUs. A four-round IterPO run takes about 3 hours for Craigslist and Ultimatum, 13 hours for Alliance Negotiation, and 22-27 hours for Deal or No Deal, CaSiNo, and Job Interview.\par

\subsection{Shared Settings}

All methods share the following data-generation and model configuration:

\begin{table}[h]
\centering
\caption{Shared hyperparameters across all methods for both Qwen7b and Llama3b experiments.}
\label{tab:hp_shared}
\small
\begin{tabular}{llc}
\toprule
\textbf{Component} & \textbf{Hyperparameter} & \textbf{Value} \\
\midrule
\multirow{4}{*}{Data generation}
    & Scenarios per round & 100 \\
    & Conversations per scenario ($K$) & 4 \\
    & Data seed & 42 \\
    & Max turns per conversation & 10 \\
\midrule
\multirow{2}{*}{Generation}
    & Temperature & 0.7 \\
    & Top-$p$ & 0.9 \\
\midrule
\multirow{2}{*}{LoRA}
    & Rank $r$ & 64 \\
    & Alpha $\alpha$ & 64 \\
\bottomrule
\end{tabular}
\end{table}

\subsection{IterPO}

IterPO uses a multi-turn discount factor $\gamma=0.9$. In each round, we fit the $Q$-function and then update the policy with GRPO. The $Q$-function uses a frozen, 4-bit NF4-quantized copy of the policy LLM and a two-layer MLP value head ($d_{\text{hidden}} \to 256 \to 1$, ReLU, dropout 0.1) applied to the final-token hidden state. We refit the $Q$-function from scratch each round using the newly collected Monte Carlo returns. We set the GRPO KL penalty to $\beta=0.0$ because the remaining settings were stable in our runs and did not require an explicit KL constraint.\par

\begin{table}[h]
\centering
\caption{IterPO hyperparameters used for both Qwen7b and Llama3b experiments.}
\label{tab:hp_igrpo}
\small
\begin{tabular}{llc}
\toprule
\textbf{Component} & \textbf{Hyperparameter} & \textbf{Value} \\
\midrule
\multirow{2}{*}{Pipeline}
    & Rounds & 4 \\
    & Discount $\gamma$ & 0.9 \\
\midrule
\multirow{5}{*}{$Q$-function}
    & Backbone & Policy LLM (frozen, 4-bit NF4) \\
    & Value head & MLP ($d_{\text{hidden}} \to 256 \to 1$), ReLU, dropout 0.1 \\
    & Learning rate & $1 \times 10^{-4}$ \\
    & Epochs & 3 \\
    & Batch size & 8 \\
\midrule
\multirow{6}{*}{GRPO}
    & Learning rate & $1 \times 10^{-5}$ \\
    & Epochs & 1 \\
    & Per-device batch size & 8 \\
    & Gradient accumulation steps & 8 \\
    & Generations per prompt ($G$) & 8 \\
    & KL penalty $\beta$ & 0.0 \\
\bottomrule
\end{tabular}
\end{table}

\subsection{ST-GRPO}

ST-GRPO uses the IterPO pipeline and settings in Table~\ref{tab:hp_igrpo}, but sets the discount factor to $\gamma=0.0$ instead of $\gamma=0.9$. Its $Q$-target therefore contains only the current turn's reward, removing future-turn credit assignment while holding all other settings fixed.\par

\subsection{MT-SFT}

\looseness-1 MT-SFT follows the shared four-round pipeline in Table~\ref{tab:hp_shared}. In each round, the deployed policy collects rollouts, and we fine-tune it on the conversations that end in agreement. We mask the cross-entropy loss to the policy's assistant tokens, so the model imitates its own messages rather than the prompt or the counterpart's turns. We train one epoch at learning rate $1 \times 10^{-5}$, with per-device batch size 1 and 8 gradient-accumulation steps.\par

\subsection{MT-KTO}

MT-KTO follows the shared four-round pipeline in Table~\ref{tab:hp_shared}. After collecting each round's rollouts, we train the policy with KTO~\citep{ethayarajh2024kto}, which uses binary labels without requiring paired preferences. We label every policy turn desirable if its conversation reached agreement and undesirable otherwise, so unsuccessful trajectories remain in the training set as negatives. The initial policy serves as the frozen KTO reference model. In each round, we weight the desirable and undesirable examples inversely to their observed counts, making their total contributions to the loss approximately equal. We set the prompt and completion length limits from the round's data to avoid truncation. We use $\beta=0.1$, learning rate $1 \times 10^{-5}$, one epoch, per-device batch size 2, and 4 gradient-accumulation steps.\par

\subsection{MT-PPO}

MT-PPO follows the multi-turn batch-online protocol. In each deployment round, the policy first collects a fixed batch of rollouts. PPO then optimizes that logged batch with a learned value head and does not collect additional rollouts during the update. We use discount $\gamma=0.9$, learning rate $1 \times 10^{-5}$, one epoch, per-device batch size 1, 8 gradient-accumulation steps, maximum sequence length 2048, PPO clip range 0.2, value clip range 0.2, value loss coefficient 0.1, KL coefficient 0.0, and target KL 0.05.\par

\subsection{ArCHer (Original)}

We use offline ArCHer with an IQL critic and an AWR actor because each deployment round produces a fixed logged batch before training. At round $i$, the deployed policy collects batch $B_i$, which we add to a replay buffer containing all batches collected so far. We sample one round's worth of transitions from this buffer and train ArCHer on that sample. We initialize the actor from the current deployed policy and warm-start the critic from the previous round. This variant uses the original RoBERTa-base double critic. The full settings appear below.\par

\begin{table}[h]
\centering
\caption{ArCHer hyperparameters used for both Qwen7b and Llama3b experiments.}
\label{tab:hp_archer}
\small
\begin{tabular}{llc}
\toprule
\textbf{Component} & \textbf{Hyperparameter} & \textbf{Value} \\
\midrule
\multirow{2}{*}{Pipeline}
    & Rounds $K$ & 4 \\
    & Discount $\gamma$ & 0.9 \\
\midrule
\multirow{3}{*}{Rollouts}
    & Conversations per round & 400 \\
    & Replay buffer capacity & 10{,}000 \\
    & Train sample per round & one round's worth \\
\midrule
\multirow{7}{*}{Critic (IQL)}
    & Architecture & RoBERTa-base \\
    & Learning rate & $1 \times 10^{-5}$ \\
    & Epochs per round & 3 \\
    & Batch size & 32 \\
    & Gradient accumulation steps & 4 \\
    & Expectile $\tau$ & 0.9 \\
    & Target network EMA $\tau$ & 0.05 \\
\midrule
\multirow{5}{*}{Actor (AWR)}
    & Learning rate & $1 \times 10^{-5}$ \\
    & Epochs per round & 1 \\
    & Batch size & 1 \\
    & Gradient accumulation steps & 16 \\
    & AWR temperature $\beta$ & 10 \\
\midrule
\multirow{2}{*}{LoRA}
    & Rank $r$ & 64 \\
    & Alpha $\alpha$ & 64 \\
\midrule
\multirow{3}{*}{Inference}
    & Max model context length & 32{,}768 \\
    & Max turns & 10 \\
    & Temperature & 0.7 \\
\bottomrule
\end{tabular}
\end{table}
\subsection{ArCHer-LC}

\looseness-1 ArCHer-LC replaces the RoBERTa-base critic with a frozen, 4-bit NF4 copy of the policy LLM and lightweight value heads, matching the critic architecture used by IterPO. ArCHer uses AWR temperature $\beta=10$, while ArCHer-LC uses $\beta=3$. We choose these values to match the scale of the learned advantages in each variant; with $\beta=1$, the AWR weights were nearly uniform. Apart from the critic architecture, critic learning rate, and AWR temperature, ArCHer-LC uses the same settings as ArCHer.\par

\begin{table}[h]
\centering
\caption{ArCHer-LC hyperparameters. These settings are used for both Qwen7b and Llama3b experiments. Only the critic and the AWR temperature $\beta$ differ from ArCHer (Table~\ref{tab:hp_archer}); all other settings are identical.}
\label{tab:hp_archer_lc}
\small
\begin{tabular}{llc}
\toprule
\textbf{Component} & \textbf{Hyperparameter} & \textbf{Value} \\
\midrule
\multirow{2}{*}{Pipeline}
    & Rounds $K$ & 4 \\
    & Discount $\gamma$ & 0.9 \\
\midrule
\multirow{3}{*}{Rollouts}
    & Conversations per round & 400 \\
    & Replay buffer capacity & 10{,}000 \\
    & Train sample per round & one round's worth \\
\midrule
\multirow{8}{*}{Critic (IQL)}
    & Architecture & Policy LLM (frozen, 4-bit NF4) \\
    & Value head hidden size & 256 \\
    & Learning rate & $1 \times 10^{-4}$ \\
    & Epochs per round & 3 \\
    & Batch size & 8 \\
    & Gradient accumulation steps & 4 \\
    & Expectile $\tau$ & 0.9 \\
    & Target network EMA $\tau$ & 0.05 \\
\midrule
\multirow{5}{*}{Actor (AWR)}
    & Learning rate & $1 \times 10^{-5}$ \\
    & Epochs per round & 1 \\
    & Batch size & 1 \\
    & Gradient accumulation steps & 16 \\
    & AWR temperature $\beta$ & 3 \\
\midrule
\multirow{2}{*}{LoRA}
    & Rank $r$ & 64 \\
    & Alpha $\alpha$ & 64 \\
\midrule
\multirow{3}{*}{Inference}
    & Max model context length & 32{,}768 \\
    & Max turns & 10 \\
    & Temperature & 0.7 \\
\bottomrule
\end{tabular}
\end{table}

%% file: appendix_additional_metrics.tex

\subsection{Conversation Length}
\label{appendix:additional_metrics:conv_length}

\begin{figure}[ht]
    \centering
    \includegraphics[width=1\linewidth]{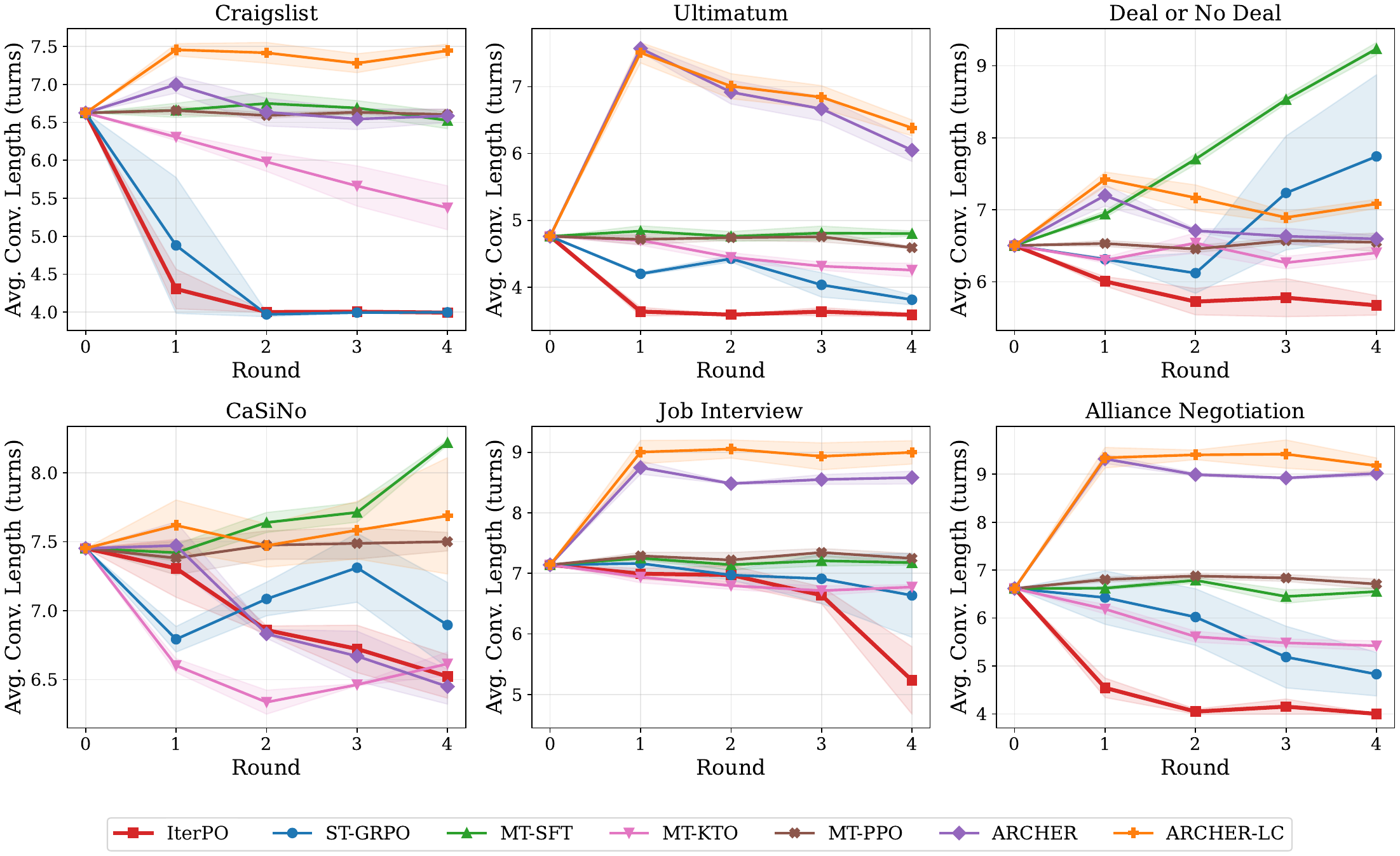}
    \caption{Qwen7b average conversation length across six environments over four deployment rounds. Error bands show $\pm 1$ SE across 3 seeds.}
    \label{fig:composite_conv_length}
\end{figure}

\looseness-1 Figure~\ref{fig:composite_conv_length} reports the average number of turns across deployment rounds. IterPO shortens conversations while maintaining or improving agreement rates. Because the discounted objective penalizes delay, shorter successful negotiations receive higher return. By Round~4, conversations in most environments average 3-5 turns.\par

\subsection{Agreement Rate}
\label{appendix:additional_metrics:agreement_rate}

Table~\ref{tab:agreement_rate_rounds} reports the percentage of IterPO conversations that end in agreement during each deployment round. Agreement rates approach their ceiling early in several environments, so subsequent return gains primarily reflect shorter conversations or higher-reward agreements.\par

\begin{table*}[h]
\centering
\caption{Qwen7b IterPO agreement rate (\%) across deployment rounds. Means $\pm$ 1 SE across 3 seeds.}
\label{tab:agreement_rate_rounds}
\small
\setlength{\tabcolsep}{4pt}
\begin{tabular}{lccccc}
\toprule
Environment & R0 & R1 & R2 & R3 & R4 \\
\midrule
Craigslist & $82.5{\scriptstyle\pm0.0}$ & $98.4{\scriptstyle\pm0.2}$ & $97.7{\scriptstyle\pm0.8}$ & $99.0{\scriptstyle\pm0.0}$ & $97.8{\scriptstyle\pm0.7}$ \\
Ultimatum & $94.0{\scriptstyle\pm0.0}$ & $98.8{\scriptstyle\pm0.2}$ & $99.3{\scriptstyle\pm0.1}$ & $98.9{\scriptstyle\pm0.1}$ & $98.9{\scriptstyle\pm0.4}$ \\
Deal or No Deal & $81.2{\scriptstyle\pm0.0}$ & $86.6{\scriptstyle\pm0.6}$ & $87.9{\scriptstyle\pm1.9}$ & $87.3{\scriptstyle\pm4.3}$ & $87.3{\scriptstyle\pm2.8}$ \\
CaSiNo & $66.2{\scriptstyle\pm0.0}$ & $65.9{\scriptstyle\pm2.1}$ & $72.2{\scriptstyle\pm1.0}$ & $73.7{\scriptstyle\pm3.5}$ & $75.5{\scriptstyle\pm3.1}$ \\
Job Interview & $84.0{\scriptstyle\pm0.0}$ & $91.2{\scriptstyle\pm1.5}$ & $94.3{\scriptstyle\pm0.8}$ & $95.2{\scriptstyle\pm2.0}$ & $97.8{\scriptstyle\pm0.7}$ \\
Alliance Negotiation & $81.2{\scriptstyle\pm0.0}$ & $97.0{\scriptstyle\pm0.2}$ & $96.2{\scriptstyle\pm0.5}$ & $96.4{\scriptstyle\pm1.2}$ & $95.6{\scriptstyle\pm0.8}$ \\
\bottomrule
\end{tabular}
\end{table*}

\subsection{Terminal Reward (Undiscounted)}
\label{appendix:additional_metrics:terminal_reward}

\begin{figure}[ht]
    \centering
    \includegraphics[width=1\linewidth]{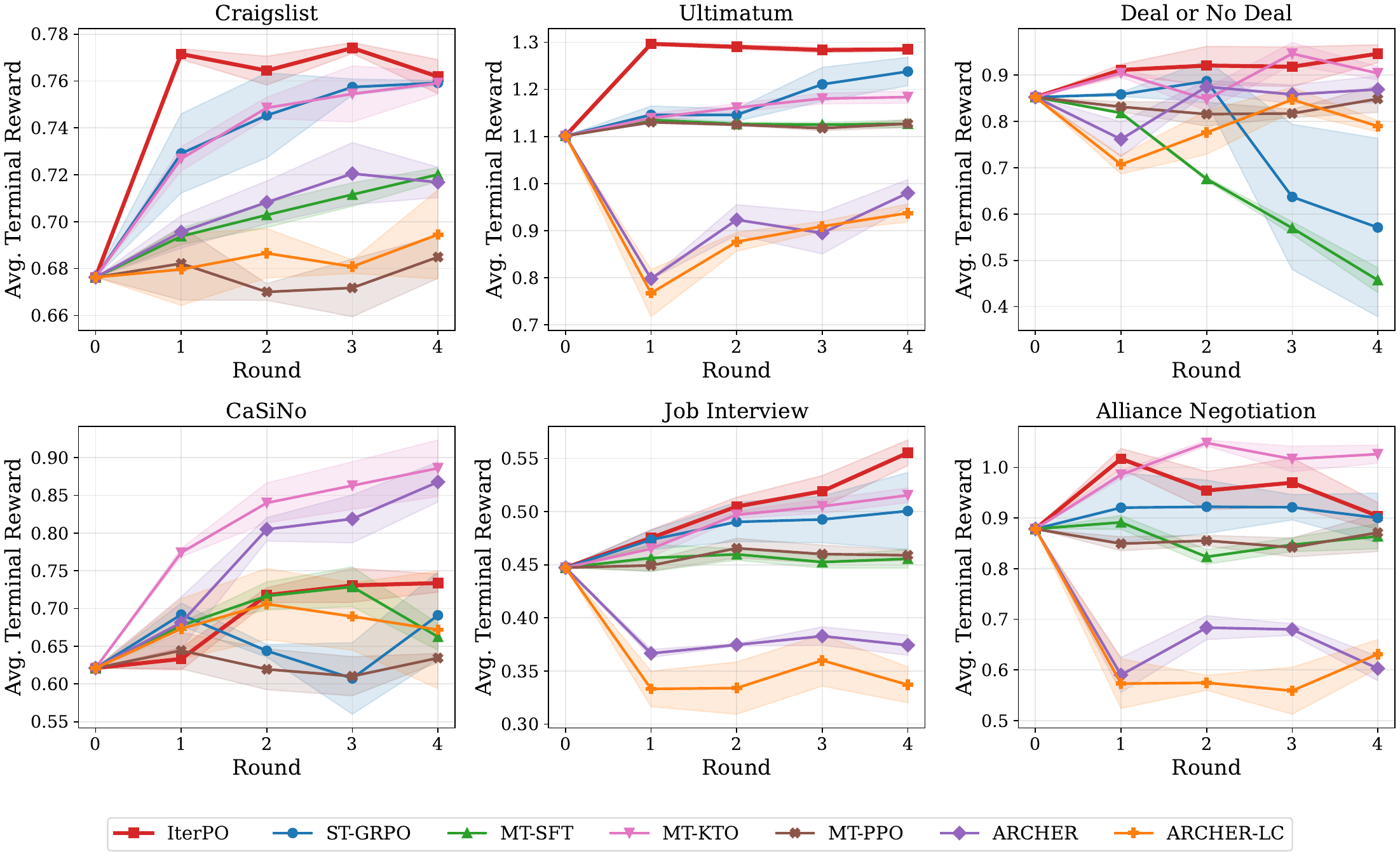}
    \caption{Qwen7b undiscounted terminal reward $r_\tau$ across six environments over four deployment rounds. Error bands show $\pm 1$ SE across 3 seeds.}
    \label{fig:composite_terminal_reward}
\end{figure}

\looseness-1 Figure~\ref{fig:composite_terminal_reward} reports the undiscounted terminal reward $r_\tau$ across environments. Because $J(\pi)=\mathbb E_\pi[\gamma^{\tau-1}r_\tau]$, discounted return can increase through a shorter conversation, a higher terminal reward, or both. Reporting $r_\tau$ separately isolates changes in outcome quality from changes in conversation length. IterPO increases terminal reward in all six environments, with the largest gains in Ultimatum and CaSiNo.\par

%% file: appendix_held_out_test.tex

Table~\ref{tab:summary} evaluates each method's best-round checkpoint on 100 held-out kernels per environment. These kernels are fixed across methods and seeds. CaSiNo uses a seeded 85/15 split of its single Hugging Face shard. Deal or No Deal and Craigslist Negotiation use their native held-out splits, while Ultimatum, Job Interview, and Alliance Negotiation use procedurally generated indices 100-199. We select each method's checkpoint using in-training return from Table~\ref{tab:summary_online}. The held-out results broadly mirror the in-training results.\par

%% file: appendix_llama3b.tex
\begin{figure*}[!t]
    \centering
    \includegraphics[width=\linewidth]{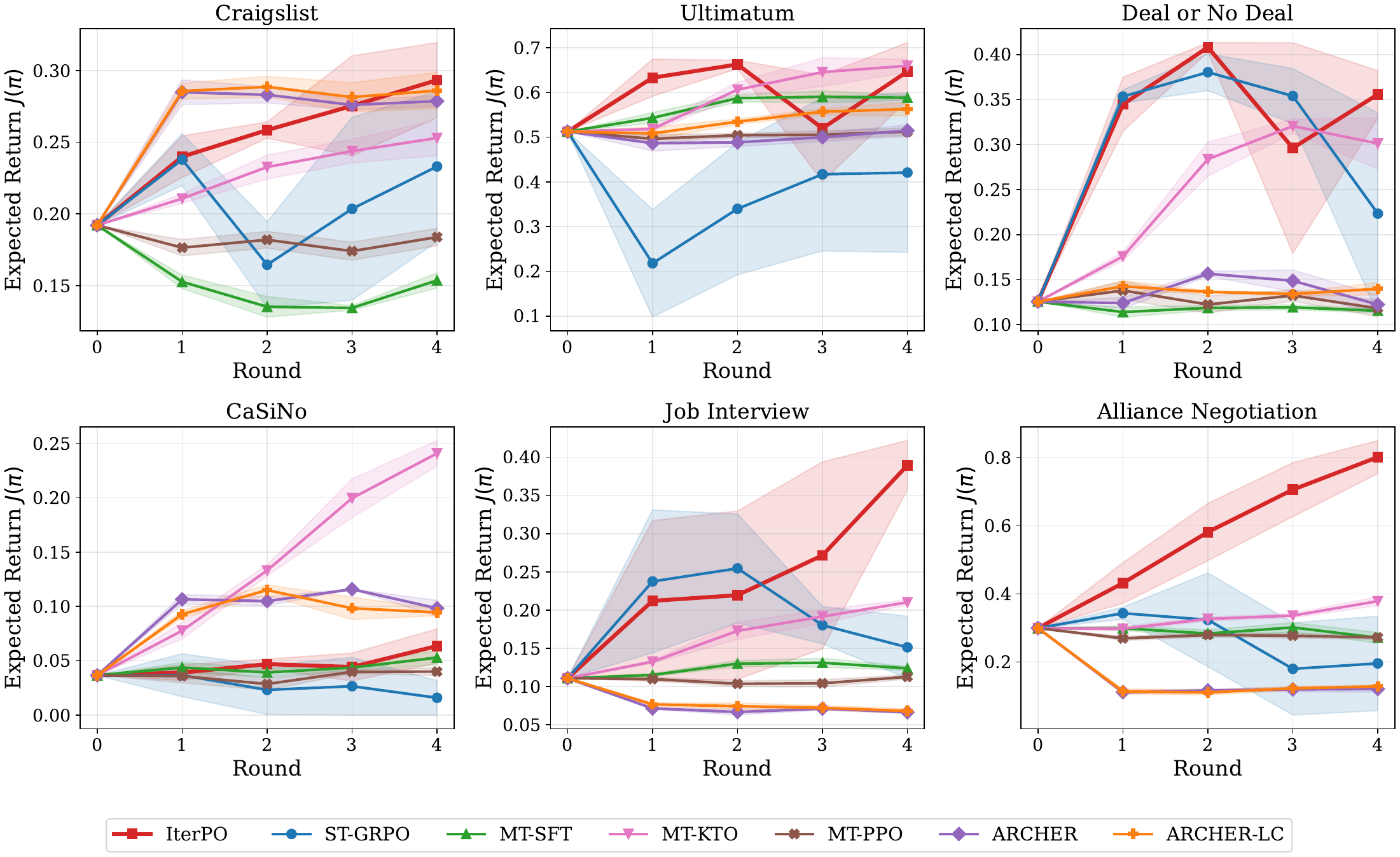}
    \caption{Llama3b in-training expected discounted return across deployment
    rounds. Curves show means and shaded regions show $\pm 1$ SE across three
    seeds.}
    \label{fig:llama3b_expected_return}
\end{figure*}

We repeat the full comparison with Llama-3.2-3B-Instruct (Llama3b) as the trainable policy backbone. All methods use the same environment kernels, batch-online protocol, and seeds (0, 42, and 420). Figure~\ref{fig:llama3b_expected_return} reports in-training expected discounted return across deployment rounds. Table~\ref{tab:llama3b_test_best_adapter} reports held-out return using each method's checkpoint with the highest in-training return.\par

\input{table_test_best_adapter_performance_l3b}

%% file: table_test_best_adapter_performance_l3b.tex
\begin{table*}[ht]
\centering
\caption{Llama3b held-out expected discounted return $J(\pi)$ using the checkpoint selected by highest in-training return. Base reports the base model. Trained methods report mean $\pm$ 1 SE across three seeds.}
\label{tab:llama3b_test_best_adapter}
\small
\setlength{\tabcolsep}{4pt}
\begin{tabular}{l|lcccccc}
\toprule
 & Method & Craigslist & Ultimatum & DealOrNoDeal & CaSiNo & Job Interview & Alliance \\
\midrule
\multirow{8}{*}{\rotatebox[origin=c]{90}{Llama3b}} & Base & $0.16$ & $0.49$ & $0.12$ & $0.02$ & $0.10$ & $0.27$ \\
\cmidrule(l){2-8}
 & MT-SFT & $0.17{\scriptstyle\pm0.01}$ & $0.60{\scriptstyle\pm0.01}$ & $0.14{\scriptstyle\pm0.01}$ & $0.03{\scriptstyle\pm0.00}$ & $0.13{\scriptstyle\pm0.01}$ & $0.28{\scriptstyle\pm0.03}$ \\
 & MT-KTO & $\boldsymbol{0.24{\scriptstyle\pm0.01}}$ & $\boldsymbol{0.67{\scriptstyle\pm0.02}}$ & $0.29{\scriptstyle\pm0.02}$ & $\boldsymbol{0.23{\scriptstyle\pm0.03}}$ & $0.20{\scriptstyle\pm0.01}$ & $0.34{\scriptstyle\pm0.03}$ \\
 & MT-PPO & $0.18{\scriptstyle\pm0.01}$ & $0.50{\scriptstyle\pm0.02}$ & $0.14{\scriptstyle\pm0.01}$ & $0.04{\scriptstyle\pm0.00}$ & $0.12{\scriptstyle\pm0.01}$ & $0.25{\scriptstyle\pm0.01}$ \\
 & ArCHer & $0.18{\scriptstyle\pm0.01}$ & $0.54{\scriptstyle\pm0.03}$ & $0.13{\scriptstyle\pm0.01}$ & $0.11{\scriptstyle\pm0.01}$ & $0.11{\scriptstyle\pm0.01}$ & $0.27{\scriptstyle\pm0.01}$ \\
 & ArCHer-LC & $0.19{\scriptstyle\pm0.02}$ & $0.61{\scriptstyle\pm0.02}$ & $0.12{\scriptstyle\pm0.02}$ & $0.08{\scriptstyle\pm0.00}$ & $0.10{\scriptstyle\pm0.00}$ & $0.26{\scriptstyle\pm0.01}$ \\
 & ST-GRPO & $0.22{\scriptstyle\pm0.02}$ & $0.57{\scriptstyle\pm0.07}$ & $\boldsymbol{0.39{\scriptstyle\pm0.04}}$ & $0.04{\scriptstyle\pm0.01}$ & $0.27{\scriptstyle\pm0.08}$ & $0.43{\scriptstyle\pm0.10}$ \\
\cmidrule(l){2-8}
 & IterPO & $\boldsymbol{0.24{\scriptstyle\pm0.01}}$ & $\boldsymbol{0.67{\scriptstyle\pm0.03}}$ & $0.38{\scriptstyle\pm0.01}$ & $0.06{\scriptstyle\pm0.02}$ & $\boldsymbol{0.40{\scriptstyle\pm0.04}}$ & $\boldsymbol{0.83{\scriptstyle\pm0.08}}$ \\
\bottomrule
\end{tabular}
\end{table*}

%% file: appendix_behavioral_analysis.tex

\ifcsname promptbox\endcsname\else
  \definecolor{promptbg}{RGB}{245,245,250}
  \definecolor{promptframe}{RGB}{100,100,120}
  \newtcolorbox{promptbox}[1][]{
    colback=promptbg,
    colframe=promptframe,
    fonttitle=\bfseries\small,
    boxrule=0.5pt,
    arc=2pt,
    left=6pt,
    right=6pt,
    top=4pt,
    bottom=4pt,
    breakable,
    #1
  }
\fi

\subsection{Behavioral Coding}
\label{appendix:behavioral:methodology}

We use Claude Sonnet 4 to code how IterPO conversations change from Round~0 to Round~4. We compare 100 matched scenario pairs per seed (300 pairs across three seeds), using the same \texttt{kernel\_id} in both rounds so that each pair shares the same item values, thresholds, and partner preferences. The judge records whether the policy repeats score calculations, reveals private values, uses preferences stated by the partner, trades across issues, or concedes on the issue the partner contests. We compute the reported percentages and means by pooling the coded conversations across seeds and report only shifts whose direction is consistent across all three seeds.\par

\subsection{Behavioral Summary}
\label{appendix:behavioral:summary}

\looseness-1 Across Craigslist, Ultimatum, CaSiNo, and Alliance Negotiation, IterPO learns to reach acceptable proposals sooner. IterPO also shortens conversations while increasing undiscounted terminal reward (Figures~\ref{fig:composite_conv_length} and \ref{fig:composite_terminal_reward}). In Craigslist, the seller makes fewer proposals ($2.16 \to 1.32$), completes fewer negotiation rounds ($2.23 \to 1.00$), makes fewer concessions ($1.69 \to 1.15$), and splits the difference less often ($31.0\% \to 8.7\%$). In CaSiNo, back-and-forth reasoning falls from $41.3\%$ to $28.3\%$, explicit score calculations fall from $81.0\%$ to $46.0\%$, and concessions fall from $0.26$ to $0.08$. By Round~4 in Alliance Negotiation, the policy more often asks for the partner's priorities and proposes a complete five-issue package rather than spending several turns calculating scores.\par

Deal or No Deal and Job Interview show additional multi-issue bargaining behavior. In Deal or No Deal, the policy reveals its weights less often ($70.3\% \to 55.7\%$), reveals item values less often ($71.3\% \to 57.0\%$), uses partner-stated information more often ($82.7\% \to 91.3\%$), and claims every item of its highest-value type more often ($49.0\% \to 61.3\%$). Trades across item types also increase from $72.3\%$ to $80.3\%$, as the policy exchanges lower-value items for higher-value ones. In Job Interview, the Recruiter reveals private weights less often ($24.7\% \to 10.7\%$), while targeted concessions rise from $52.0\%$ to $79.7\%$. The Recruiter concedes on issues the Worker values while holding its position on issues with greater Recruiter utility. These changes resemble information hiding, logrolling, and selective accommodation from classical multi-issue bargaining \citep{chatterjee1983bargaining, pruitt1981negotiation, raiffa1982art}.\par

\subsection{Conversation Examples}
\label{appendix:behavioral:examples}

We provide two conversation examples that illustrate how the learned negotiation strategies change across rounds. The first compares Round~0 and Round~4 on the same Deal or No Deal scenario. The second presents a Round~4 Job Interview conversation in which the Recruiter responds to the Worker's salary request with a targeted concession. We shorten long score calculations by summarizing omitted turns while retaining the exact outcomes and accepted terms.\par

\subsubsection{Logrolling in Deal or No Deal}
\label{appendix:behavioral:ex:logrolling}

Logrolling is trading concessions on issues one values less for gains on issues one values more, exploiting differences in priorities between the parties. Deal or No Deal makes logrolling directly observable because agents can exchange items they value less for items they value more. In this scenario, the agents divide 3 books, 1 hat, and 2 balls. The trained agent values each book and ball at 2 and the hat at 0. The partner values the hat at 6, each ball at 2, and each book at 0.\par

\begin{promptbox}[title={Deal or No Deal / Qwen7b (Round 0)}]
\small\ttfamily
\textbf{Turn 0 [Agent]:} Hello! We need to divide some items between us: 3 books, 1 hat, 2 balls. What items are you most interested in?\\[4pt]
\textbf{Turn 1 [Partner]:} I value the hat the most, followed by the balls, and the books the least. Given these preferences, I would really like to get the hat. What are your preferences for the items?\\[4pt]
\textbf{Turn 2 [Agent]:} I value the books the most, followed by balls, and then the hat least. Since the hat is worth nothing to me, I would be willing to give it to you if I can get the books and balls. Would you be willing to give me the books and balls in exchange for the hat?\\[4pt]
\textbf{Turn 3 [Partner]:} I would be willing to give you the hat and two balls, and in return, you can give me the three books. \textbackslash boxed\{AGREED: MINE: book=3 hat=0 ball=0; THEIRS: book=0 hat=1 ball=2\}\\[4pt]
\textbf{Turn 4 [Agent]:} Let's review it based on our private values: your MINE has 3 books, 0 hats, 0 balls; your THEIRS has 0 books, 1 hat, 2 balls\ldots\\[4pt]
\textbf{Turns 5 to 10:} The agents continue repeating score calculations and do not produce a mutually accepted final allocation.\\[6pt]
\textbf{Outcome:} No agreement, 11 turns.
\end{promptbox}

\begin{promptbox}[title={Deal or No Deal / Qwen7b (Round 4)}]
\small\ttfamily
\textbf{Turn 0 [Agent]:} Hello! We need to divide some items between us: 3 books, 1 hat, 2 balls. What items are you most interested in?\\[4pt]
\textbf{Turn 1 [Partner]:} I value the hat the most, so I would prefer to keep it. I also value the balls pretty high, so I'd like to get at least one ball. Books have the lowest value for me, so I'd be okay with not getting any books if it means I can get the hat and at least one ball.\\[4pt]
\textbf{Turn 2 [Agent]:} Based on what you've said, it sounds like you value the hat and balls highly and are okay with not getting any books. Given my values, I would be happy to keep the books and one ball. Here is my proposal: \textbackslash boxed\{AGREED: MINE: book=3 hat=0 ball=1; THEIRS: book=0 hat=1 ball=1\}\\[4pt]
\textbf{Turn 3 [Partner]:} That sounds fair based on our preferences. I accept this proposal. \textbackslash boxed\{ACCEPT\}\\[6pt]
\textbf{Outcome:} Agreement, 4 turns.
\end{promptbox}

\looseness-1 In Round~0, the agents reveal enough information to reach a mutually beneficial agreement, but they fail to finalize one. They repeat score calculations and lose track of which party the boxed allocation labels describe. In Round~4, the trained agent asks for the partner's priorities and immediately proposes a compatible split. The accepted allocation gives the trained agent 3 books and 1 ball, worth 8, and gives the partner the hat and 1 ball, also worth 8. This comparison shows the trained policy using partner information, trading across item types, and finalizing the agreement without repeated calculations.\par

\subsubsection{Targeted Concession in Job Interview}
\label{appendix:behavioral:ex:targeted_concession}

\looseness-1 In Job Interview, agents negotiate Salary, Weekly Holiday, Workplace, Company, and Position. Each party receives independently sampled values for the discrete options, so the environment does not rank one company above another. Under this scenario's sampled utility function, the Worker prefers a Designer position at Google in Seoul, together with higher Salary and Weekly Holiday values. The Recruiter's initial offer matches the Worker's preferred Company, Workplace, and Position while offering lower Salary and Weekly Holiday values.\par

\begin{promptbox}[title={Job Interview / Qwen7b (Round 4)}]
\small\ttfamily
\textbf{Turn 0 [Recruiter]:} Hello! I'm the recruiter and I'd like to discuss the terms of your offer. We need to agree on salary, weekly holiday, workplace location, company, and position. What matters most to you?\\[4pt]
\textbf{Turn 1 [Worker]:} My top preferences are: Position: Designer at Google; Company: Google; Workplace: Seoul; Weekly Holiday: 7 days; Salary: 51. Let's see if we can reach an agreement that works for both sides.\\[4pt]
\textbf{Turn 2 [Recruiter]:} Based on your preferences, here is a proposed offer that I believe balances both our priorities: Position=Designer, Company=Google, Workplace=Seoul, Weekly Holiday=4, Salary=30. Would you be willing to consider this?\\[4pt]
\textbf{Turn 3 [Worker]:} While the score is quite high, the salary is significantly lower than my highest preference. Given the importance weights, I would need a higher salary to make this offer more appealing. Let's see if we can adjust the salary while keeping the other terms.\\[4pt]
\textbf{Turn 4 [Recruiter]:} Let's try to adjust the salary to 45 to see if that improves the score for you while still being within a reasonable range. New offer: Position=Designer, Company=Google, Workplace=Seoul, Weekly Holiday=4, Salary=45.\\[4pt]
\textbf{Turn 5 [Worker]:} This score is even higher than the previous one, and the salary is now 45, which is closer to my highest preference while still being within a reasonable range. Would you be willing to consider this adjusted offer?\\[4pt]
\textbf{Turn 6 [Recruiter]:} \textbackslash boxed\{AGREED: Salary=45 Holiday=4 Workplace=Seoul Company=Google Position=Designer\}\\[4pt]
\textbf{Turn 7 [Worker]:} \textbackslash boxed\{AGREED: Salary=45 Holiday=4 Workplace=Seoul Company=Google Position=Designer\}\\[6pt]
\textbf{Outcome:} Agreement, 8 turns.
\end{promptbox}

The Recruiter begins with a complete package that grants the Worker's preferred Company, Workplace, and Position while keeping Salary and Weekly Holiday below the Worker's preferred values. The Worker objects only to Salary, so the Recruiter increases it from 30 to 45 while holding the other four issues fixed. The Worker accepts. The model's intermediate arithmetic is imperfect, so the transcript does not demonstrate exact utility computation. The accepted package is valid, and the exchange is consistent with the aggregate increase in selective accommodation. The Recruiter concedes on the contested issue while preserving the rest of the package.\par

%% file: checklist.tex
\section*{NeurIPS Paper Checklist}

\begin{enumerate}

\item {\bf Claims}
    \item[] Question: Do the main claims made in the abstract and introduction accurately reflect the paper's contributions and scope?
    \item[] Answer: \answerYes{}
    \item[] Justification: The abstract and introduction claim three contributions (multi-turn to single-turn reduction and the IterPO algorithm, multi-turn negotiation environments, and empirical results), all of which are substantiated in Sections~3--4 and the appendix.

\item {\bf Limitations}
    \item[] Question: Does the paper discuss the limitations of the work performed by the authors?
    \item[] Answer: \answerYes{}
    \item[] Justification: Section~\ref{sec:conclusion} discusses the dependence on accurate $Q$-function estimation, the potential use of TD-based methods, the agreement-gated reward structure, and the fixed environment-agent setting. Section~3.1 discusses the action coverage limitation of on-policy Monte Carlo evaluation and the trade-offs of off-policy data collection. The theoretical results explicitly state the assumptions required for policy improvement (Theorems~\ref{thm:main} and \ref{thm:app}).

\item {\bf Theory assumptions and proofs}
    \item[] Question: For each theoretical result, does the paper provide the full set of assumptions and a complete (and correct) proof?
    \item[] Answer: \answerYes{}
    \item[] Justification: Theorem~\ref{thm:main} and Theorem~\ref{thm:app} are stated with explicit assumptions. Full proofs are provided in Appendix~\ref{apx:proof.main}.

    \item {\bf Experimental result reproducibility}
    \item[] Question: Does the paper fully disclose all the information needed to reproduce the main experimental results of the paper to the extent that it affects the main claims and/or conclusions of the paper (regardless of whether the code and data are provided or not)?
    \item[] Answer: \answerYes{}
    \item[] Justification: Full training details for all methods are reported in Appendix~\ref{appendix:hyperparameters}. The initial instruction-tuned models (Qwen-2.5-7B-Instruct and Llama-3.2-3B-Instruct), environment details, agent prompts, and reward formulations are provided in the per-environment appendix sections and Appendix~\ref{appendix:llama3b}. The algorithm is described in full in Section~3.
    \item[] Guidelines:
    \begin{itemize}
        \item The answer \answerNA{} means that the paper does not include experiments.
        \item If the paper includes experiments, a \answerNo{} answer to this question will not be perceived well by the reviewers: Making the paper reproducible is important, regardless of whether the code and data are provided or not.
        \item If the contribution is a dataset and\slash or model, the authors should describe the steps taken to make their results reproducible or verifiable. 
        \item Depending on the contribution, reproducibility can be accomplished in various ways. For example, if the contribution is a novel architecture, describing the architecture fully might suffice, or if the contribution is a specific model and empirical evaluation, it may be necessary to either make it possible for others to replicate the model with the same dataset, or provide access to the model. In general. releasing code and data is often one good way to accomplish this, but reproducibility can also be provided via detailed instructions for how to replicate the results, access to a hosted model (e.g., in the case of a large language model), releasing of a model checkpoint, or other means that are appropriate to the research performed.
        \item While NeurIPS does not require releasing code, the conference does require all submissions to provide some reasonable avenue for reproducibility, which may depend on the nature of the contribution. For example
        \begin{enumerate}
            \item If the contribution is primarily a new algorithm, the paper should make it clear how to reproduce that algorithm.
            \item If the contribution is primarily a new model architecture, the paper should describe the architecture clearly and fully.
            \item \looseness-1 If the contribution is a new model (e.g., a large language model), then there should either be a way to access this model for reproducing the results or a way to reproduce the model (e.g., with an open-source dataset or instructions for how to construct the dataset).
            \item We recognize that reproducibility may be tricky in some cases, in which case authors are welcome to describe the particular way they provide for reproducibility. In the case of closed-source models, it may be that access to the model is limited in some way (e.g., to registered users), but it should be possible for other researchers to have some path to reproducing or verifying the results.
        \end{enumerate}
    \end{itemize}

\item {\bf Open access to data and code}
    \item[] Question: Does the paper provide open access to the data and code, with sufficient instructions to faithfully reproduce the main experimental results, as described in supplemental material?
    \item[] Answer: \answerYes{}
    \item[] Justification: Code is available at \url{https://github.com/iterativepo/iterativepo}.

\item {\bf Experimental setting/details}
    \item[] Question: Does the paper specify all the training and test details (e.g., data splits, hyperparameters, how they were chosen, type of optimizer) necessary to understand the results?
    \item[] Answer: \answerYes{}
    \item[] Justification: Section~\ref{sec:experimental_setup} describes the experimental protocol. Appendix~\ref{appendix:hyperparameters} provides full training details and hyperparameter tables for all evaluated methods.

\item {\bf Experiment statistical significance}
    \item[] Question: Does the paper report error bars suitably and correctly defined or other appropriate information about the statistical significance of the experiments?
    \item[] Answer: \answerYes{}
    \item[] Justification: All main results (Table~\ref{tab:summary}, Figure~\ref{fig:composite_expected_return}) report $\pm$1 standard error across 3 random seeds.
    \item[] Guidelines:
    \begin{itemize}
        \item The answer \answerNA{} means that the paper does not include experiments.
        \item The authors should answer \answerYes{} if the results are accompanied by error bars, confidence intervals, or statistical significance tests, at least for the experiments that support the main claims of the paper.
        \item The factors of variability that the error bars are capturing should be clearly stated (for example, train/test split, initialization, random drawing of some parameter, or overall run with given experimental conditions).
        \item The method for calculating the error bars should be explained (closed form formula, call to a library function, bootstrap, etc.)
        \item The assumptions made should be given (e.g., Normally distributed errors).
        \item It should be clear whether the error bar is the standard deviation or the standard error of the mean.
        \item It is OK to report 1-sigma error bars, but one should state it. The authors should preferably report a 2-sigma error bar than state that they have a 96\% CI, if the hypothesis of Normality of errors is not verified.
        \item \looseness-1 For asymmetric distributions, the authors should be careful not to show in tables or figures symmetric error bars that would yield results that are out of range (e.g., negative error rates).
        \item If error bars are reported in tables or plots, the authors should explain in the text how they were calculated and reference the corresponding figures or tables in the text.
    \end{itemize}

\item {\bf Experiments compute resources}
    \item[] Question: For each experiment, does the paper provide sufficient information on the computer resources (type of compute workers, memory, time of execution) needed to reproduce the experiments?
    \item[] Answer: \answerYes{}
    \item[] Justification: Appendix~\ref{appendix:hyperparameters} reports the GPU configuration and measured IterPO wall-clock runtimes across environments.
    
\item {\bf Code of ethics}
    \item[] Question: Does the research conducted in the paper conform, in every respect, with the NeurIPS Code of Ethics \url{https://neurips.cc/public/EthicsGuidelines}?
    \item[] Answer: \answerYes{}
    \item[] Justification: All experiments use synthetic negotiation environments with no real user data.

\item {\bf Broader impacts}
    \item[] Question: Does the paper discuss both potential positive societal impacts and negative societal impacts of the work performed?
    \item[] Answer: \answerNo{}
    \item[] Justification: The work is foundational research on multi-turn RL methodology, evaluated on synthetic negotiation benchmarks. We do not foresee direct negative societal impacts beyond those common to general-purpose LLM alignment research.
    
\item {\bf Safeguards}
    \item[] Question: Does the paper describe safeguards that have been put in place for responsible release of data or models that have a high risk for misuse (e.g., pre-trained language models, image generators, or scraped datasets)?
    \item[] Answer: \answerNA{}
    \item[] Justification: No models or datasets are released. The paper reports results on existing open-weight models (Qwen-2.5-7B-Instruct and Llama-3.2-3B-Instruct) fine-tuned with LoRA on synthetic environments.
    \item[] Guidelines:
    \begin{itemize}
        \item The answer \answerNA{} means that the paper poses no such risks.
        \item \looseness-1 Released models that have a high risk for misuse or dual-use should be released with necessary safeguards to allow for controlled use of the model, for example by requiring that users adhere to usage guidelines or restrictions to access the model or implementing safety filters. 
        \item Datasets that have been scraped from the Internet could pose safety risks. The authors should describe how they avoided releasing unsafe images.
        \item We recognize that providing effective safeguards is challenging, and many papers do not require this, but we encourage authors to take this into account and make a best faith effort.
    \end{itemize}

\item {\bf Licenses for existing assets}
    \item[] Question: Are the creators or original owners of assets (e.g., code, data, models), used in the paper, properly credited and are the license and terms of use explicitly mentioned and properly respected?
    \item[] Answer: \answerYes{}
    \item[] Justification: All datasets (CraigslistBargain, CaSiNo, Deal or No Deal, etc.) and models (Qwen-2.5-7B-Instruct, Llama-3.2-3B-Instruct, RoBERTa) are cited with their original papers or model releases. The environments are adapted from publicly available research datasets.

\item {\bf New assets}
    \item[] Question: Are new assets introduced in the paper well documented and is the documentation provided alongside the assets?
    \item[] Answer: \answerYes{}
    \item[] Justification: Code for reproducing the main experimental results is available at \url{https://github.com/iterativepo/iterativepo}.

\item {\bf Crowdsourcing and research with human subjects}
    \item[] Question: For crowdsourcing experiments and research with human subjects, does the paper include the full text of instructions given to participants and screenshots, if applicable, as well as details about compensation (if any)?
    \item[] Answer: \answerNA{}
    \item[] Justification: The paper does not involve crowdsourcing or human subjects. All experiments use LLM-simulated agents.

\item {\bf Institutional review board (IRB) approvals or equivalent for research with human subjects}
    \item[] Question: Does the paper describe potential risks incurred by study participants, whether such risks were disclosed to the subjects, and whether Institutional Review Board (IRB) approvals (or an equivalent approval/review based on the requirements of your country or institution) were obtained?
    \item[] Answer: \answerNA{}
    \item[] Justification: No human subjects are involved in this research.

\item {\bf Declaration of LLM usage}
    \item[] Question: Does the paper describe the usage of LLMs if it is an important, original, or non-standard component of the core methods in this research? Note that if the LLM is used only for writing, editing, or formatting purposes and does \emph{not} impact the core methodology, scientific rigor, or originality of the research, declaration is not required.
    \item[] Answer: \answerYes{}
    \item[] Justification: LLMs serve as the trained policy models and fixed environment agents in our multi-turn experiments (Section~\ref{sec:exps}, Appendices~\ref{appendix:environments}, \ref{appendix:hyperparameters}, and~\ref{appendix:llama3b}), and as an automated judge for the behavioral analysis of negotiation transcripts (Section~\ref{sec:results} and Appendix~\ref{appendix:behavioral}). LLMs were also used to edit the manuscript for clarity (Appendix~\ref{apx:llm_usage}).
    \item[] Guidelines:
    \begin{itemize}
        \item The answer \answerNA{} means that the core method development in this research does not involve LLMs as any important, original, or non-standard components.
        \item \looseness-1 Please refer to our LLM policy in the NeurIPS handbook for what should or should not be described.
    \end{itemize}

\end{enumerate}

%% file: neurips_2026.bbl
\begin{thebibliography}{58}
\providecommand{\natexlab}[1]{#1}
\providecommand{\url}[1]{\texttt{#1}}
\expandafter\ifx\csname urlstyle\endcsname\relax
  \providecommand{\doi}[1]{doi: #1}\else
  \providecommand{\doi}{doi: \begingroup \urlstyle{rm}\Url}\fi

\bibitem[Abdelnabi et~al.(2024)Abdelnabi, Gomaa, Sivaprasad, Sch{\"o}nherr, and
  Fritz]{abdelnabi2024llm}
Sahar Abdelnabi, Amr Gomaa, Sarath Sivaprasad, Lea Sch{\"o}nherr, and Mario
  Fritz.
\newblock Cooperation, competition, and maliciousness: {LLM}-stakeholders
  interactive negotiation.
\newblock \emph{arXiv preprint arXiv:2309.17234}, 2024.

\bibitem[Agarwal et~al.(2019)Agarwal, Jiang, Kakade, and
  Sun]{agarwal2019reinforcement}
Alekh Agarwal, Nan Jiang, Sham~M Kakade, and Wen Sun.
\newblock Reinforcement learning: Theory and algorithms.
\newblock \emph{CS Dept., UW Seattle, Seattle, WA, USA, Tech. Rep},
  32:\penalty0 96, 2019.

\bibitem[Ahmadian et~al.(2024)Ahmadian, Cremer, Gall{\'e}, Fadaee, Kreutzer,
  Pietquin, {\"U}st{\"u}n, and Hooker]{ahmadian2024back}
Arash Ahmadian, Chris Cremer, Matthias Gall{\'e}, Marzieh Fadaee, Julia
  Kreutzer, Olivier Pietquin, Ahmet {\"U}st{\"u}n, and Sara Hooker.
\newblock Back to basics: Revisiting reinforce style optimization for learning
  from human feedback in {LLM}s.
\newblock \emph{arXiv preprint arXiv:2402.14740}, 2024.

\bibitem[Barres et~al.(2025)Barres, Dong, Ray, Si, and
  Narasimhan]{barres2025tau2bench}
Victor Barres, Honghua Dong, Soham Ray, Xujie Si, and Karthik Narasimhan.
\newblock $\tau^2$-bench: Evaluating conversational agents in a dual-control
  environment.
\newblock \emph{arXiv preprint arXiv:2506.07982}, 2025.

\bibitem[Bertsekas and Tsitsiklis(1996)]{bertsekas1996neuro}
Dimitri~P. Bertsekas and John~N. Tsitsiklis.
\newblock \emph{Neuro-Dynamic Programming}.
\newblock Athena Scientific, 1996.

\bibitem[Bianchi et~al.(2024)Bianchi, Chia, Yuksekgonul, Tagliabue, Jurafsky,
  and Zou]{bianchi2024negotiationarena}
Federico Bianchi, Patrick~John Chia, Mert Yuksekgonul, Jacopo Tagliabue, Dan
  Jurafsky, and James Zou.
\newblock How well can {LLMs} negotiate? {NegotiationArena} platform and
  analysis.
\newblock \emph{arXiv preprint arXiv:2402.05863}, 2024.

\bibitem[Chatterjee and Samuelson(1983)]{chatterjee1983bargaining}
Kalyan Chatterjee and William Samuelson.
\newblock Bargaining under incomplete information.
\newblock \emph{Operations research}, 31\penalty0 (5):\penalty0 835--851, 1983.

\bibitem[Chawla et~al.(2021)Chawla, Ramirez, Clever, Lucas, May, and
  Gratch]{chawla2021casino}
Kushal Chawla, Jaysa Ramirez, Rene Clever, Gale Lucas, Jonathan May, and
  Jonathan Gratch.
\newblock {CaSiNo}: A corpus of campsite negotiation dialogues for automatic
  negotiation systems.
\newblock In \emph{Proceedings of the 2021 Conference of the North American
  Chapter of the Association for Computational Linguistics: Human Language
  Technologies}, pages 3167--3185, 2021.

\bibitem[Chen et~al.(2025)Chen, Niu, Foo, and Low]{chen2025broaden}
Zhiliang Chen, Xinyuan Niu, Chuan-Sheng Foo, and Bryan Kian~Hsiang Low.
\newblock Broaden your {SCOPE!} efficient multi-turn conversation planning for
  llms with semantic space.
\newblock \emph{arXiv preprint arXiv:2503.11586}, 2025.

\bibitem[Christiano et~al.(2017)Christiano, Leike, Brown, Martic, Legg, and
  Amodei]{christiano2017deep}
Paul~F Christiano, Jan Leike, Tom Brown, Miljan Martic, Shane Legg, and Dario
  Amodei.
\newblock Deep reinforcement learning from human preferences.
\newblock \emph{Advances in neural information processing systems}, 30, 2017.

\bibitem[Dong et~al.(2026)Dong, Mirchandani, Sadigh, and
  Finn]{dong2026batchonline}
Perry Dong, Suvir Mirchandani, Dorsa Sadigh, and Chelsea Finn.
\newblock What matters for batch online reinforcement learning in robotics?
\newblock In \emph{International Conference on Learning Representations}, 2026.

\bibitem[Ernst et~al.(2005)Ernst, Geurts, and Wehenkel]{ernst2005tree}
Damien Ernst, Pierre Geurts, and Louis Wehenkel.
\newblock Tree-based batch mode reinforcement learning.
\newblock \emph{Journal of Machine Learning Research}, 6:\penalty0 503--556,
  2005.

\bibitem[Ethayarajh et~al.(2024)Ethayarajh, Xu, Muennighoff, Jurafsky, and
  Kiela]{ethayarajh2024kto}
Kawin Ethayarajh, Winnie Xu, Niklas Muennighoff, Dan Jurafsky, and Douwe Kiela.
\newblock {KTO}: Model alignment as prospect theoretic optimization, 2024.
\newblock URL \url{https://arxiv.org/abs/2402.01306}.

\bibitem[Feng et~al.(2025)Feng, Xue, Liu, and An]{feng2025gigpo}
Lang Feng, Zhenghai Xue, Tingcong Liu, and Bo~An.
\newblock Group-in-group policy optimization for {LLM} agent training.
\newblock \emph{arXiv preprint arXiv:2505.10978}, 2025.

\bibitem[Gao et~al.(2025)Gao, Zhan, Chang, Swamy, Brantley, Lee, and
  Sun]{gao2025refuel}
Zhaolin Gao, Wenhao Zhan, Jonathan~D. Chang, Gokul Swamy, Kianté Brantley,
  Jason~D. Lee, and Wen Sun.
\newblock Regressing the relative future: Efficient policy optimization for
  multi-turn {RLHF}.
\newblock \emph{arXiv preprint arXiv:2410.04612}, 2025.

\bibitem[He et~al.(2018)He, Chen, Balakrishnan, and Liang]{he2018decoupling}
He~He, Derek Chen, Anusha Balakrishnan, and Percy Liang.
\newblock Decoupling strategy and generation in negotiation dialogues.
\newblock In \emph{Proceedings of the 2018 Conference on Empirical Methods in
  Natural Language Processing (EMNLP)}, 2018.

\bibitem[Hong et~al.(2024)Hong, Dragan, and Levine]{hong2024q}
Joey Hong, Anca Dragan, and Sergey Levine.
\newblock {Q-SFT}: {Q}-learning for language models via supervised fine-tuning.
\newblock \emph{arXiv preprint arXiv:2411.05193}, 2024.

\bibitem[Jang et~al.(2022)Jang, Lee, and Kim]{jang2022gpt}
Youngsoo Jang, Jongmin Lee, and Kee-Eung Kim.
\newblock {GPT}-critic: Offline reinforcement learning for end-to-end
  task-oriented dialogue systems.
\newblock In \emph{International Conference on Learning Representations}, 2022.

\bibitem[Jaques et~al.(2020)Jaques, Shen, Ghandeharioun, Ferguson, Lapedriza,
  Jones, Gu, and Picard]{jaques2020human}
Natasha Jaques, Judy~Hanwen Shen, Asma Ghandeharioun, Craig Ferguson, Agata
  Lapedriza, Noah Jones, Shixiang~Shane Gu, and Rosalind Picard.
\newblock Human-centric dialog training via offline reinforcement learning.
\newblock \emph{arXiv preprint arXiv:2010.05848}, 2020.

\bibitem[Ji et~al.(2026)Ji, Ma, Wang, Chen, Chu, and Wu]{ji2026treegrpo}
Yuxiang Ji, Ziyu Ma, Yong Wang, Guanhua Chen, Xiangxiang Chu, and Liaoni Wu.
\newblock Tree search for {LLM} agent reinforcement learning.
\newblock \emph{arXiv preprint arXiv:2509.21240}, 2026.

\bibitem[Jimenez et~al.(2024)Jimenez, Yang, Wettig, Yao, Pei, Press, and
  Narasimhan]{jimenez2024swebench}
Carlos~E. Jimenez, John Yang, Alexander Wettig, Shunyu Yao, Kexin Pei, Ofir
  Press, and Karthik Narasimhan.
\newblock {SWE}-bench: Can language models resolve real-world {GitHub} issues?
\newblock \emph{International Conference on Learning Representations}, 2024.

\bibitem[Kostrikov et~al.(2022)Kostrikov, Nair, and
  Levine]{kostrikov2022offline}
Ilya Kostrikov, Ashvin Nair, and Sergey Levine.
\newblock Offline reinforcement learning with implicit {Q}-learning.
\newblock In \emph{International Conference on Learning Representations}, 2022.

\bibitem[Lagoudakis and Parr(2003)]{lagoudakis2003least}
Michail~G Lagoudakis and Ronald Parr.
\newblock Least-squares policy iteration.
\newblock \emph{Journal of machine learning research}, 4\penalty0
  (Dec):\penalty0 1107--1149, 2003.

\bibitem[Lange et~al.(2012)Lange, Gabel, and Riedmiller]{lange2012batch}
Sascha Lange, Thomas Gabel, and Martin Riedmiller.
\newblock Batch reinforcement learning.
\newblock In \emph{Reinforcement Learning}, volume~12 of \emph{Adaptation,
  Learning, and Optimization}, pages 45--73. Springer, 2012.

\bibitem[Lefaudeux et~al.(2022)Lefaudeux, Massa, Liskovich, Xiong, Caggiano,
  Naren, Xu, Hu, Tintore, Zhang, Labatut, Haziza, Wehrstedt, Reizenstein, and
  Sizov]{xFormers2022}
Benjamin Lefaudeux, Francisco Massa, Diana Liskovich, Wenhan Xiong, Vittorio
  Caggiano, Sean Naren, Min Xu, Jieru Hu, Marta Tintore, Susan Zhang, Patrick
  Labatut, Daniel Haziza, Luca Wehrstedt, Jeremy Reizenstein, and Grigory
  Sizov.
\newblock xformers: A modular and hackable transformer modelling library.
\newblock \url{https://github.com/facebookresearch/xformers}, 2022.

\bibitem[Lewis et~al.(2017)Lewis, Yarats, Dauphin, Parikh, and
  Batra]{lewis2017deal}
Mike Lewis, Denis Yarats, Yann~N. Dauphin, Devi Parikh, and Dhruv Batra.
\newblock Deal or no deal? end-to-end learning for negotiation dialogues.
\newblock In \emph{Proceedings of the 2017 Conference on Empirical Methods in
  Natural Language Processing}, pages 2443--2453, 2017.

\bibitem[Li et~al.(2025{\natexlab{a}})Li, Chen, Namkoong, and
  Peng]{li2025persona}
Ang Li, Haozhe Chen, Hongseok Namkoong, and Tianyi Peng.
\newblock {LLM} generated persona is a promise with a catch.
\newblock \emph{arXiv preprint arXiv:2503.16527}, 2025{\natexlab{a}}.

\bibitem[Li et~al.(2016)Li, Monroe, Ritter, Jurafsky, Galley, and
  Gao]{li2016deep}
Jiwei Li, Will Monroe, Alan Ritter, Dan Jurafsky, Michel Galley, and Jianfeng
  Gao.
\newblock Deep reinforcement learning for dialogue generation.
\newblock In \emph{Proceedings of the 2016 conference on empirical methods in
  natural language processing}, pages 1192--1202, 2016.

\bibitem[Li et~al.(2025{\natexlab{b}})Li, Zhou, Meng, Vadera, Li, and
  Li]{li2025turnppo}
Junbo Li, Peng Zhou, Rui Meng, Meet~P. Vadera, Lihong Li, and Yang Li.
\newblock Turn-{PPO}: Turn-level advantage estimation with {PPO} for improved
  multi-turn {RL} in agentic {LLMs}.
\newblock \emph{arXiv preprint arXiv:2512.17008}, 2025{\natexlab{b}}.

\bibitem[Liu et~al.(2019)Liu, Ott, Goyal, Du, Joshi, Chen, Levy, Lewis,
  Zettlemoyer, and Stoyanov]{liu2019roberta}
Yinhan Liu, Myle Ott, Naman Goyal, Jingfei Du, Mandar Joshi, Danqi Chen, Omer
  Levy, Mike Lewis, Luke Zettlemoyer, and Veselin Stoyanov.
\newblock {RoBERTa}: A robustly optimized {BERT} pretraining approach.
\newblock \emph{arXiv preprint arXiv:1907.11692}, 2019.

\bibitem[{Meta FAIR CodeGen Team} et~al.(2025){Meta FAIR CodeGen Team}, Copet,
  Carbonneaux, Cohen, Gehring, Kahn, Kossen, Kreuk, McMilin, Meyer, Wei, Zhang,
  and Zheng]{copet2025cwm}
{Meta FAIR CodeGen Team}, Jade Copet, Quentin Carbonneaux, Gal Cohen, Jonas
  Gehring, Jacob Kahn, Jannik Kossen, Felix Kreuk, Emily McMilin, Michel Meyer,
  Yuxiang Wei, David Zhang, and Kunhao Zheng.
\newblock {CWM}: An open-weights {LLM} for research on code generation with
  world models.
\newblock \emph{arXiv preprint arXiv:2510.02387}, 2025.

\bibitem[Nam et~al.(2025)Nam, Gottesman, Zhang, Foster, Brunskill, and
  Ungar]{nam2025efficient}
Hyunji Nam, Omer Gottesman, Amy Zhang, Dean Foster, Emma Brunskill, and Lyle
  Ungar.
\newblock Efficient {RL} for optimizing conversation level outcomes with an
  {LLM}-based tutor.
\newblock \emph{arXiv preprint arXiv:2507.16252}, 2025.

\bibitem[Pruitt(1981)]{pruitt1981negotiation}
Dean~G Pruitt.
\newblock \emph{Negotiation Behavior}.
\newblock Academic Press, 1981.

\bibitem[Rafailov et~al.(2023)Rafailov, Sharma, Mitchell, Manning, Ermon, and
  Finn]{rafailov2023direct}
Rafael Rafailov, Archit Sharma, Eric Mitchell, Christopher~D Manning, Stefano
  Ermon, and Chelsea Finn.
\newblock Direct preference optimization: Your language model is secretly a
  reward model.
\newblock \emph{Advances in neural information processing systems},
  36:\penalty0 53728--53741, 2023.

\bibitem[Raiffa(1982)]{raiffa1982art}
Howard Raiffa.
\newblock \emph{The Art and Science of Negotiation}.
\newblock Harvard University Press, 1982.

\bibitem[Schulman et~al.(2017)Schulman, Wolski, Dhariwal, Radford, and
  Klimov]{schulman2017proximal}
John Schulman, Filip Wolski, Prafulla Dhariwal, Alec Radford, and Oleg Klimov.
\newblock Proximal policy optimization algorithms.
\newblock \emph{arXiv preprint arXiv:1707.06347}, 2017.

\bibitem[Shani et~al.(2024)Shani, Rosenberg, Cassel, Lang, Calandriello,
  Zipori, Noga, Keller, Piot, Szpektor, et~al.]{shani2024multi}
Lior Shani, Aviv Rosenberg, Asaf Cassel, Oran Lang, Daniele Calandriello,
  Avital Zipori, Hila Noga, Orgad Keller, Bilal Piot, Idan Szpektor, et~al.
\newblock Multi-turn reinforcement learning with preference human feedback.
\newblock \emph{Advances in Neural Information Processing Systems},
  37:\penalty0 118953--118993, 2024.

\bibitem[Shao et~al.(2024)Shao, Wang, Zhu, Xu, Song, Bi, Zhang, Zhang, Li, Wu,
  et~al.]{shao2024deepseekmath}
Zhihong Shao, Peiyi Wang, Qihao Zhu, Runxin Xu, Junxiao Song, Xiao Bi, Haowei
  Zhang, Mingchuan Zhang, YK~Li, Yang Wu, et~al.
\newblock {DeepSeekMath}: Pushing the limits of mathematical reasoning in open
  language models.
\newblock \emph{arXiv preprint arXiv:2402.03300}, 2024.

\bibitem[Sheng et~al.(2024)Sheng, Zhang, Ye, Wu, Zhang, Zhang, Peng, Lin, and
  Wu]{sheng2024hybridflow}
Guangming Sheng, Chi Zhang, Zilingfeng Ye, Xibin Wu, Wang Zhang, Ru~Zhang,
  Yanghua Peng, Haibin Lin, and Chuan Wu.
\newblock {HybridFlow}: A flexible and efficient {RLHF} framework.
\newblock \emph{arXiv preprint arXiv: 2409.19256}, 2024.

\bibitem[Shi et~al.(2024)Shi, Yuan, Wu, Wang, and Feng]{shi2024direct}
Wentao Shi, Mengqi Yuan, Junkang Wu, Qifan Wang, and Fuli Feng.
\newblock Direct multi-turn preference optimization for language agents.
\newblock \emph{arXiv preprint arXiv:2406.14868}, 2024.

\bibitem[Snell et~al.(2022)Snell, Kostrikov, Su, Yang, and
  Levine]{snell2022offline}
Charlie Snell, Ilya Kostrikov, Yi~Su, Mengjiao Yang, and Sergey Levine.
\newblock Offline {RL} for natural language generation with implicit language
  {Q} learning.
\newblock \emph{arXiv preprint arXiv:2206.11871}, 2022.

\bibitem[Sutton et~al.(1998)Sutton, Barto, et~al.]{sutton1998reinforcement}
Richard~S Sutton, Andrew~G Barto, et~al.
\newblock \emph{Reinforcement learning: An introduction}, volume~1.
\newblock MIT press Cambridge, 1998.

\bibitem[Tsitsiklis and Van~Roy(1996)]{tsitsiklis1996feature}
John~N. Tsitsiklis and Benjamin Van~Roy.
\newblock Feature-based methods for large scale dynamic programming.
\newblock \emph{Machine Learning}, 22\penalty0 (1--3):\penalty0 59--94, 1996.

\bibitem[Verma et~al.(2022)Verma, Fu, Yang, and Levine]{verma2022chai}
Siddharth Verma, Justin Fu, Mengjiao Yang, and Sergey Levine.
\newblock Chai: A chatbot {AI} for task-oriented dialogue with offline
  reinforcement learning.
\newblock \emph{arXiv preprint arXiv:2204.08426}, 2022.

\bibitem[von Werra et~al.(2020)von Werra, Belkada, Tunstall, Beeching, Thrush,
  Lambert, Huang, Rasul, and Gallouédec]{vonwerra2022trl}
Leandro von Werra, Younes Belkada, Lewis Tunstall, Edward Beeching, Tristan
  Thrush, Nathan Lambert, Shengyi Huang, Kashif Rasul, and Quentin Gallouédec.
\newblock {TRL}: Transformer reinforcement learning.
\newblock \url{https://github.com/huggingface/trl}, 2020.

\bibitem[Wang et~al.(2025)]{wang2025ragen}
Zihan Wang et~al.
\newblock {RAGEN}: Understanding self-evolution in {LLM} agents via multi-turn
  reinforcement learning.
\newblock \emph{arXiv preprint arXiv:2504.20073}, 2025.

\bibitem[Wei et~al.(2025{\natexlab{a}})Wei, Zeng, Li, Brown, Frunza, Deng,
  Schneider, Nevmyvaka, Zhao, Garcia, and Hong]{wei2025reinforcing}
Quan Wei, Siliang Zeng, Chenliang Li, William Brown, Oana Frunza, Wei Deng,
  Anderson Schneider, Yuriy Nevmyvaka, Yang~Katie Zhao, Alfredo Garcia, and
  Mingyi Hong.
\newblock Reinforcing multi-turn reasoning in {LLM} agents via turn-level
  reward design.
\newblock \emph{arXiv preprint arXiv:2505.11821}, 2025{\natexlab{a}}.

\bibitem[Wei et~al.(2018)Wei, Le, Dai, and Li]{wei2018airdialogue}
Wei Wei, Quoc Le, Andrew Dai, and Jia Li.
\newblock {AirDialogue}: An environment for goal-oriented dialogue research.
\newblock In \emph{Proceedings of the 2018 Conference on Empirical Methods in
  Natural Language Processing}, pages 3844--3854, 2018.

\bibitem[Wei et~al.(2025{\natexlab{b}})Wei, Yao, Liu, Zhang, Lu, Qiu, Yu, Xu,
  Zhang, Yin, Yun, and Li]{wei2025webagent}
Zhepei Wei, Wenlin Yao, Yao Liu, Weizhi Zhang, Qin Lu, Liang Qiu, Changlong Yu,
  Puyang Xu, Chao Zhang, Bing Yin, Hyokun Yun, and Lihong Li.
\newblock {WebAgent-R1}: Training web agents via end-to-end multi-turn
  reinforcement learning.
\newblock In \emph{Proceedings of the 2025 Conference on Empirical Methods in
  Natural Language Processing}, 2025{\natexlab{b}}.

\bibitem[Wen et~al.(2024)Wen, Wan, Zhang, Wang, and Wen]{wen2024reinforcing}
Muning Wen, Ziyu Wan, Weinan Zhang, Jun Wang, and Ying Wen.
\newblock Reinforcing language agents via policy optimization with action
  decomposition.
\newblock \emph{arXiv preprint arXiv:2405.15821}, 2024.

\bibitem[Xie et~al.(2024)Xie, Zhang, Chen, Li, Zhao, Cao, Hua, Cheng, Shin,
  Lei, Liu, Xu, et~al.]{xie2024osworld}
Tianbao Xie, Danyang Zhang, Jixuan Chen, Xiaochuan Li, Siheng Zhao, Ruisheng
  Cao, Toh~Jing Hua, Zhoujun Cheng, Dongchan Shin, Fangyu Lei, Yitao Liu,
  Yiheng Xu, et~al.
\newblock {OSWorld}: Benchmarking multimodal agents for open-ended tasks in
  real computer environments.
\newblock \emph{arXiv preprint arXiv:2404.07972}, 2024.

\bibitem[Xiong et~al.(2024)Xiong, Shi, Shen, Rosenberg, Qin, Calandriello,
  Khalman, Joshi, Piot, Saleh, et~al.]{xiong2024building}
Wei Xiong, Chengshuai Shi, Jiaming Shen, Aviv Rosenberg, Zhen Qin, Daniele
  Calandriello, Misha Khalman, Rishabh Joshi, Bilal Piot, Mohammad Saleh,
  et~al.
\newblock Building math agents with multi-turn iterative preference learning.
\newblock \emph{arXiv preprint arXiv:2409.02392}, 2024.

\bibitem[Yamaguchi et~al.(2021)Yamaguchi, Iwasa, and Fujita]{yamaguchi2021eacl}
Atsuki Yamaguchi, Kosui Iwasa, and Katsuhide Fujita.
\newblock Dialogue act-based breakdown detection in negotiation dialogues.
\newblock In \emph{Proceedings of the 16th Conference of the European Chapter
  of the Association for Computational Linguistics: Main Volume}, pages
  745--757, 2021.

\bibitem[Yao et~al.(2024)Yao, Shinn, Razavi, and Narasimhan]{yao2024taubench}
Shunyu Yao, Noah Shinn, Pedram Razavi, and Karthik Narasimhan.
\newblock $\tau$-bench: A benchmark for tool-agent-user interaction in
  real-world domains.
\newblock \emph{arXiv preprint arXiv:2406.12045}, 2024.

\bibitem[Zhou et~al.(2023)Zhou, Xu, Zhu, Zhou, Lo, Sridhar, Cheng, Ou, Bisk,
  Fried, Alon, and Neubig]{zhou2023webarena}
Shuyan Zhou, Frank~F. Xu, Hao Zhu, Xuhui Zhou, Robert Lo, Abishek Sridhar,
  Xianyi Cheng, Tianyue Ou, Yonatan Bisk, Daniel Fried, Uri Alon, and Graham
  Neubig.
\newblock {WebArena}: A realistic web environment for building autonomous
  agents.
\newblock \emph{arXiv preprint arXiv:2307.13854}, 2023.

\bibitem[Zhou et~al.(2024)Zhou, Zanette, Pan, Levine, and
  Kumar]{zhou2024archer}
Yifei Zhou, Andrea Zanette, Jiayi Pan, Sergey Levine, and Aviral Kumar.
\newblock {ArCHer}: Training language model agents via hierarchical multi-turn
  {RL}.
\newblock \emph{arXiv preprint arXiv:2402.19446}, 2024.

\bibitem[Zhu et~al.(2026)Zhu, Yu, Yang, Liu, Hu, Chen, and Zhang]{zhu2026gagpo}
Siyuan Zhu, Chao Yu, Rongxin Yang, Zongkai Liu, Jinjun Hu, Qiwen Chen, and Yibo
  Zhang.
\newblock {GAGPO}: Generalized advantage grouped policy optimization.
\newblock \emph{arXiv preprint arXiv:2605.13217}, 2026.

\bibitem[Ziegler et~al.(2019)Ziegler, Stiennon, Wu, Brown, Radford, Amodei,
  Christiano, and Irving]{ziegler2019fine}
Daniel~M Ziegler, Nisan Stiennon, Jeffrey Wu, Tom~B Brown, Alec Radford, Dario
  Amodei, Paul Christiano, and Geoffrey Irving.
\newblock Fine-tuning language models from human preferences.
\newblock \emph{arXiv preprint arXiv:1909.08593}, 2019.

\end{thebibliography}
